%% file: boxer_arxiv.tex
\documentclass[10pt,twocolumn,letterpaper]{article}

\usepackage[pagenumbers]{cvpr} % To force page numbers, e.g. for an arXiv version

\usepackage{microtype,nicefrac,amsfonts,times,epsfig,graphicx,amsmath,amssymb,booktabs,comment,tabulary,multirow,xspace,rotating,url,makecell,soul,setspace,pifont,makebox,threeparttable}
\usepackage[font=small]{caption}

\usepackage[table]{xcolor}
\definecolor{mydarkblue}{rgb}{0,0.08,0.45}
\usepackage[pagebackref,
            breaklinks,
            colorlinks,
            linkcolor=mydarkblue,
            citecolor=mydarkblue,
            filecolor=mydarkblue,]{hyperref}

\newcommand{\boldparagraph}[1]{\vspace{0.05em}\noindent{\bf #1} }
\newcommand{\cmark}{\ding{51}}%
\newcommand{\xmark}{\ding{55}}%
\newcommand{\softmax}{\mathrm{softmax}\xspace}%

\def\train{{\texttt{train}}\xspace}
\def\val{{\texttt{val}}\xspace}

\def\test{{\texttt{test-dev}}\xspace}

\newcommand{\ours}[0]{BoxeR\xspace}
\newcommand{\boxattn}[0]{Box-Attention\xspace}
\newcommand{\boxattnop}[0]{BoxAttention\xspace}  % an operator is better defined without dash
\newcommand{\instanceattn}[0]{Instance-Attention\xspace}

\newlength\savewidth\newcommand\shline{\noalign{\global\savewidth\arrayrulewidth
  \global\arrayrulewidth 1pt}\hline\noalign{\global\arrayrulewidth\savewidth}}

\newcommand{\tablestyle}[2]{\setlength{\tabcolsep}{#1}\renewcommand{\arraystretch}{#2}\centering}

\makeatletter
\DeclareRobustCommand\onedot{\futurelet\@let@token\@onedot}
\def\@onedot{\ifx\@let@token.\else.\null\fi\xspace}

\def\eg{\emph{e.g}\onedot} 
\def\ie{\emph{i.e}\onedot} 
 
 \def\vs{\emph{vs}\onedot}
\def\wrt{w.r.t\onedot} 
\def\etal{\emph{et al}\onedot}
\makeatother

\usepackage[capitalize]{cleveref}
\crefname{section}{Sec.}{Secs.}
\Crefname{section}{Section}{Sections}
\Crefname{table}{Table}{Tables}
\crefname{table}{Tab.}{Tabs.}

\def\cvprPaperID{4311} % *** Enter the CVPR Paper ID here
\def\confName{CVPR}
\def\confYear{2022}

\begin{document}

%\title{\boxattn for Vision Transformers}
\title{\ours: \boxattn for 2D and 3D Transformers} %suggest by cs

% \author{Duy-Kien Nguyen$^{1}$~~~~~Jihong Ju$^{2}$~~~~~Olaf Booji$^{2}$~~~~~Martin R. Oswald$^{1}$~~~~~Cees G. M. Snoek$^{1}$ \\ [1mm]
% {\normalsize $^1$Atlas Lab, University of Amsterdam~~~~~$^2$TomTom}\\
% {\tt\small \{d.k.nguyen, m.r.oswald, cgmsnoek\}@uva.nl~~~~~\{jihong.ju, olaf.booji\}@tomtom.com}
% }
\author{Duy-Kien Nguyen$^{1}$~~~~~Jihong Ju$^{2}$~~~~~Olaf Booij$^{2}$~~~~~Martin R. Oswald$^{1}$~~~~~Cees G. M. Snoek$^{1}$ \\
% {\normalsize Atlas Lab} \\
{\normalsize Atlas Lab - $^1$University of Amsterdam~~~~~$^2$TomTom}\\
{\tt\small \{d.k.nguyen, m.r.oswald, cgmsnoek\}@uva.nl~~~~~\{jihong.ju, olaf.booij\}@tomtom.com}
}
% \and
\vspace{-0.2in}

\maketitle

%%%%%%%%% ABSTRACT
\begin{abstract}
In this paper, we propose a simple attention mechanism, we call \boxattn. It enables spatial interaction between grid features, as sampled from boxes of interest, and improves the learning capability of transformers for several vision tasks. Specifically, we present \ours, short for Box Transformer, which attends to a set of boxes by predicting their transformation from a reference window on an input feature map. The \ours computes attention weights on these boxes by considering its grid structure. Notably, \ours-2D naturally reasons about box information within its attention module, making it suitable for end-to-end instance detection and segmentation tasks. By learning invariance to rotation in the box-attention module, \ours-3D is capable of generating discriminative information from a bird's-eye view plane for 3D end-to-end object detection. Our experiments demonstrate that the proposed \ours-2D achieves state-of-the-art results on COCO detection and instance segmentation. Besides, \ours-3D improves over the end-to-end 3D object detection baseline and already obtains a compelling performance for the vehicle category of Waymo Open, without any class-specific optimization. Code is available at \url{https://github.com/kienduynguyen/BoxeR}.
\end{abstract}

%%%%%%%%% BODY TEXT
\section{Introduction}
\label{sec:intro}

\input{introduction}

\begin{figure*}[t]
\centering
\includegraphics[width=0.9\linewidth]{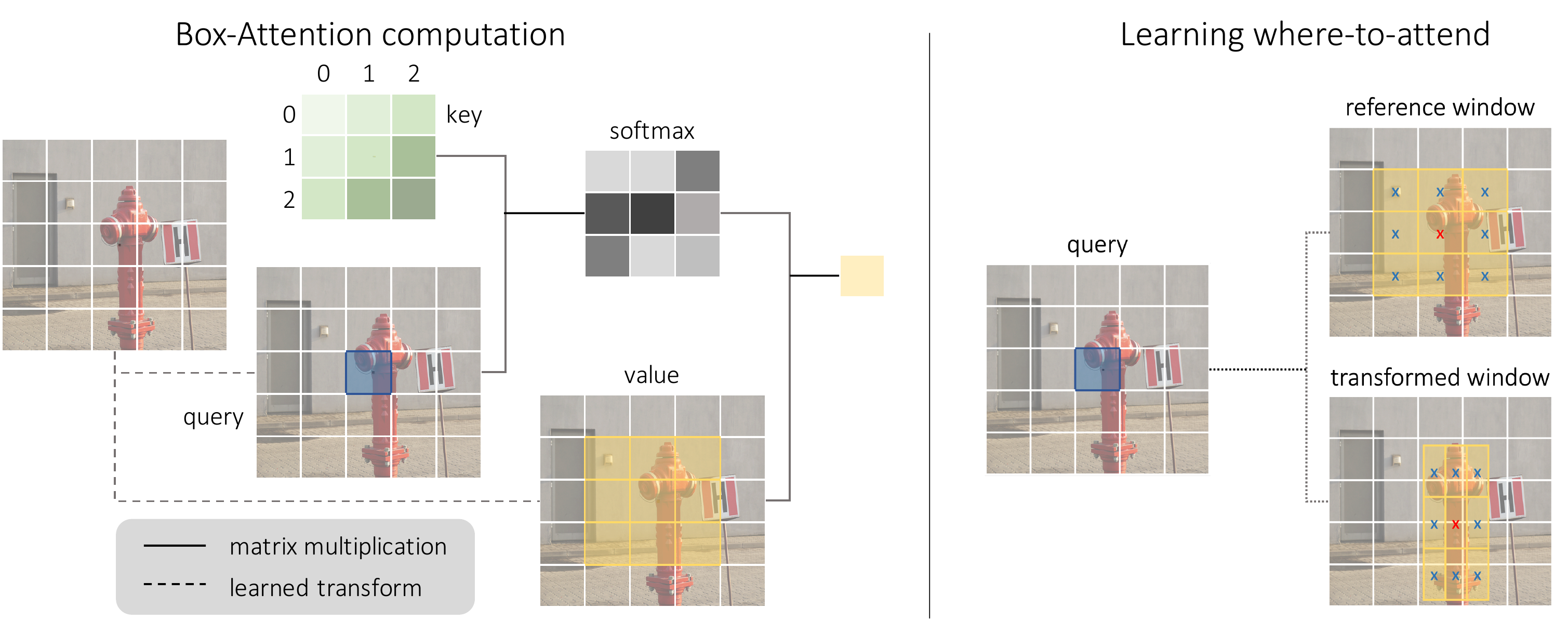}\\[-7pt]
\caption{
\textbf{\boxattn.} Left: attention computation in \boxattn with a reference window (denoted in yellow) without any transformation. Given a query vector, the \boxattn computes an attention map over $3 {\times} 3$ grid features with the query position as its center. The attention weights are generated by a matrix multiplication between query and relative position encodings as key. Right: \boxattn without and with transformations. The \boxattn with transformations is able to focus on a dynamic region in the image.}
\label{fig:box_attn}
\vspace{-0.2in}
\end{figure*}

% ------------------------------------------------------------------------
\section{Related Work}

\input{related_work}

%-------------------------------------------------------------------------
\section{\boxattn}
\label{sec:boxattention}

\input{box_attention}

%-------------------------------------------
\section{\ours-2D: A Box Transformer}
\label{sec:2Dboxer}

\input{boxer2d}

% -------------------------------------------------------------------------

\section{\ours-3D: End-to-end 3D Object Detection}
\label{sec:3Dboxer}

\input{boxer3d}

% -------------------------------------------------------------------------

\section{Experiments}
\label{sec:experiments}

\input{experiments}
\section{Conclusion and Limitations}

In this paper, we presented a transformer-based detector for end-to-end object detection and instance segmentation named \ours. The core of the network is the box-attention, which is designed to attend to an image region by learning the transformations from an initial reference window. Because of its flexibility, \ours can perform both 2D and 3D end-to-end object detection along with instance segmentation without hand-crafted modules. Experiments on the COCO and Waymo Open datasets confirm the effectiveness of the proposed architecture. 

Similar to other transformer-based architectures, we observed a larger memory footprint during the training of our networks compared to convolution-based architectures such as Faster R-CNN or Mask R-CNN. This results in the need of more advanced GPUs and higher energy consumption. Moreover, under the same FLOPs, our box-attention is slower than a convolution operation. The reasons may come from the unordered memory access of the grid sampling in our box-attention and the highly-optimized hardware and implementation for the traditional convolution. We expect to mitigate some of these problems with a more optimized implementation.

\section*{Acknowledgements}

This work has been financially supported by TomTom, the University of Amsterdam and the allowance of Top consortia for Knowledge and Innovation (TKIs) from the Netherlands Ministry of Economic Affairs and Climate Policy.

\appendix

\input{supplementary}

% \clearpage
%%%%%%%%% REFERENCES
{\small
\bibliographystyle{ieee_fullname}
\bibliography{boxer}
}

\end{document}

%% file: introduction.tex
For object detection, instance segmentation, image classification and many other current computer vision challenges, it may seem a transformer with multi-head self-attention is all one needs \cite{vaswani2017transformer}. After its success in natural language processing, learning long range feature dependencies has proven an effective tactic in computer vision too, \eg, \cite{nicolas2020detr,dosovitskiy2021vit}. Surprisingly, existing transformers for computer vision do not explicitly consider the inherent regularities of the vision modality. Importantly, the image features are vectorized in exactly the same way as language tokens, resulting in the loss of local connectivity among pixels. Once fed with sufficient data, a traditional transformer may be powerful enough to compensate for this loss of spatial structure, but in this paper we rather prefer to equip the transformer with spatial image-awareness by design. Recent evidence \cite{tay2021comparison,ascoli2021convit,zhu2020deformable} reveals that an inductive bias is of crucial importance in both natural language processing and computer vision, and the leading works on image recognition \cite{liu2021swintransformer} and object detection \cite{zhu2020deformable} all utilize ``spatial information''. Furthermore, a strong and effective inductive bias enables us to converge faster and generalize better \cite{tay2021comparison}.
% \cs{why, what is the benefit?} \kien{1. a strong inductive bias gives a better performance even in a setting of more data shown in this \url{https://arxiv.org/abs/2105.03322}; 2. it is more generalized to other tasks?; 3. reduce the computational requirement}
%By doing so it can focus on learning the task at hand, rather than figuring out which pixels are most important.

\begin{figure}[t]
\centering
\includegraphics[width=0.9\linewidth]{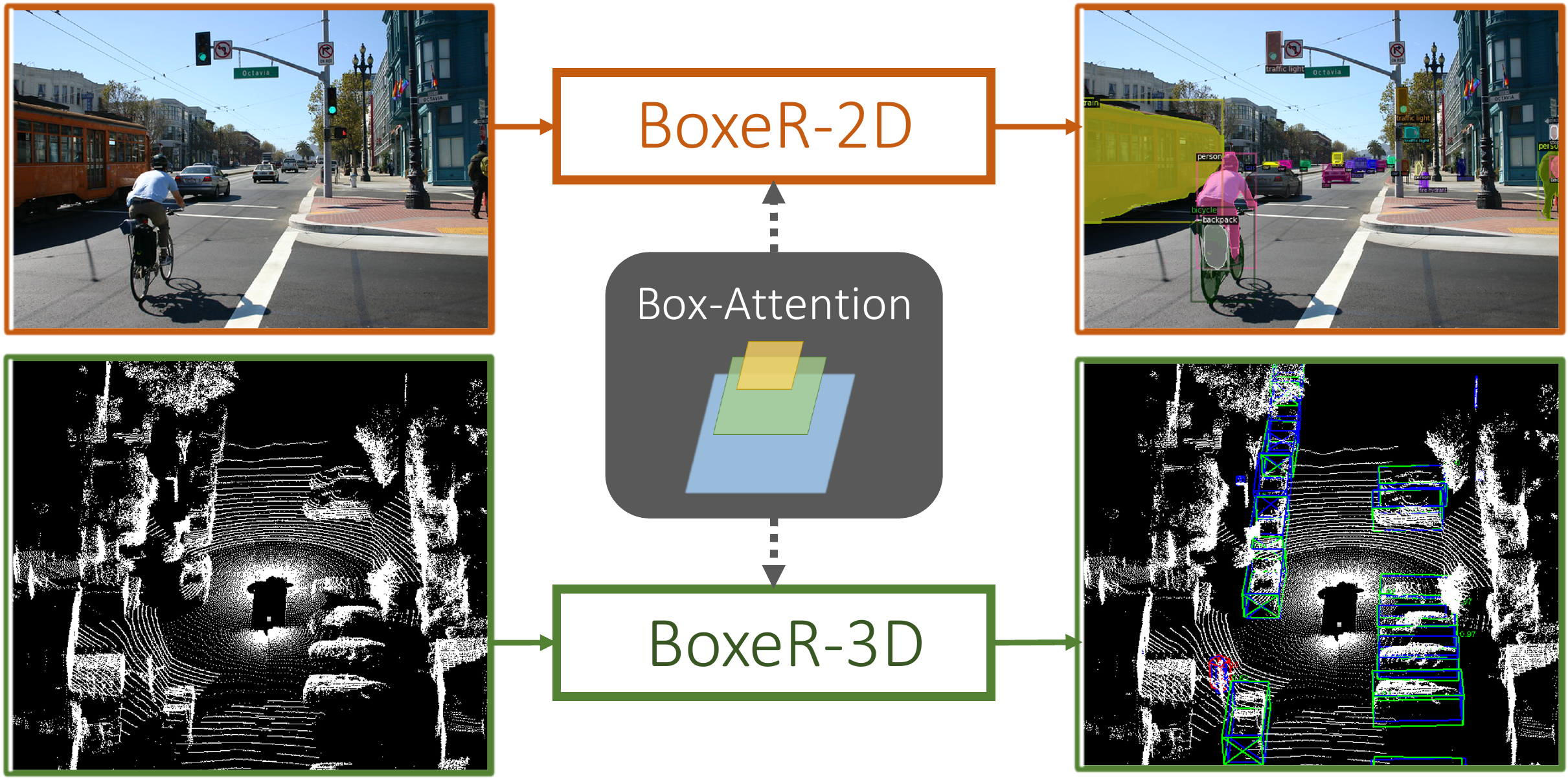}\\[-6pt]
\caption{
\textbf{\ours with box-attention} for object detection and instance segmentation. \ours-2D perceives an image and generates object bounding boxes and pixel masks. Extended from \ours-2D, \ours-3D predicts 3D bounding boxes from point cloud input.}
\label{fig:teaser}
\vspace{-0.25in}
\end{figure}

A solution is to enrich image features with positional encoding, which explicitly encodes the position information at the feature level. This is already common practice when applying multi-head attention to vision tasks. Both Carion~\etal~\cite{nicolas2020detr} and Zhu~\etal \cite{zhu2020deformable} convert absolute 2D positions, while Ramachandran \etal \cite{prajit2019visionattention} encode relative 2D positions into vectors and sum them up to image features in the attention computation. However, this approach only acts as a data augmentation to image features. It requires the network to infer the spatial information implicitly inside its weight, causing a slow convergence rate during training due to the lack of spatial-awareness in the network architecture. 
It is well known that an inductive bias in the network architecture delivers a strong ability to learn, which has been proven by well-known architectures such as the convolutional neural network \cite{lecun95convolutional} and the long short-term memory \cite{hochreiter1997lstm}. In particular, we postulate a better spatial inductive bias in the transformer's attention module leads to a better learned representation of image features.

% \cs{Paragraph 2: Inspiration from other ML advances that have profited from including inductive biasses from vision. Most notably the ConvNet that exploits convolution. Add also example from LSTM, RL, maybe others? The conclusion is that we also include a strong inductive bias from vision into a powerful ML-framework, the transformer.}
%CS: this CVPR paper is relevant for paragraph 2:
%https://arxiv.org/abs/1605.02971
%Inserts Gabor filters in first layer of CNN, so you no longer need to learn the first layer filters.

%\cs{We only contribute box attention? That is a bit too little for a conference like ICCV. Needs to be written wiser.}
% \cs{I would like to see here three explicit contributions that we claim in this paper.}
Motivated by this observation, the first contribution of this paper is a {\it \boxattn} mechanism for end-to-end vision representation learning using transformers that we present in Section~\ref{sec:boxattention}. 
% Given a query vector from image features, it first generates a box on the feature map as a region of interest for each attention head. The attention module then computes attention maps on a grid of vectors sampled from the region to generate the attended feature. 
Instead of using image features within a region of interest, it treats a set of learnable embeddings representing relative positions in the grid structure as the key vectors in the attention computation. In our second contribution, in Section~\ref{sec:2Dboxer}, these computations are encapsulated into a composite network that we call \ours-2D, short for Box transformeR, which enables a better prediction in end-to-end object detection and instance segmentation tasks. In Section~\ref{sec:3Dboxer}, the \ours-2D and box-attention are then extended into \ours-3D to tackle end-to-end 3D object detection without the requirements for 3D-IoU computation, anchors, and a heatmap of object centers. This extension to 3D object detection serves as our third contribution, see Fig. \ref{fig:teaser}.

In Section~\ref{sec:experiments}, we show the effectiveness of our contributions by several experimental results on the COCO dataset \cite{lin2014mscoco}, achieving leading results in end-to-end object detection. The proposed method introduces a simple solution for end-to-end instance segmentation that outperforms many well-established and highly-optimized architectures with fewer number of parameters on the challenging COCO instance segmentation dataset.
By utilizing only data-independent prior information, our method presents a compelling solution for end-to-end 3D object detection on the Waymo Open dataset \cite{sun2020waymodata}.

%% file: related_work.tex
We briefly review recent developments in computer vision with focus on attention mechanisms for backbones, object detection, instance segmentation and 3D object detection. 

\boldparagraph{Attention for Vision Backbones.} With the advancement of attention mechanisms, there are several approaches to create and use attention in convolutional networks, \eg,~\cite{max2015spatialtransformer,prajit2019visionattention,wang2018nonlocal,srinivas2021bottleneck}. It was recently shown in the Vision Transformer (ViT) \cite{dosovitskiy2021vit} that an attention-only network achieves comparable performance in image recognition, and outperforms convolutional neural networks in the setting of more data and longer training time.
As the ViT becomes computationally more expensive with high resolution images, while only producing a single-scale feature map, several works \cite{liu2021swintransformer,fan2021mstransformer} have focused on speeding up the self-attention computation and generating multi-scale feature maps for object detection and segmentation. 
In this paper, we instead focus on the prediction module which takes features extracted from vision backbones as inputs and provides a prediction for several vision tasks.

% \subsection{Attention for vision applications}

% The common framework of existing methods in many computer vision problems is to extract features from an input image using a backbone, and extracted features are fed into specialized modules at a later step to predict results.

\boldparagraph{Attention for Object Detection.} Modern two-stage object detection methods \cite{liu2020survay} (\ie, Faster R-CNN \cite{ren2015faster_rcnn}) utilize a region proposal network (RPN) and a prediction module on top of a pretrained backbone to predict a set of pre-defined objects. The attention mechanism is then considered as an addition of the RPN and prediction modules to further improve performance in \cite{dai2021dynamic_head,sun2021sparse_rcnn}. Alternatively, one-stage object detection methods \cite{redmon2016yolo,wang2021yolov4} remove the need for RPN and predict objects directly from convolutional feature maps. While the detection performance improves considerably, these convolution-based architectures still rely on many hand-crafted components. Recently, Carion~\etal introduced a transformer-based prediction model, called DETR~\cite{nicolas2020detr}, which gave the prediction in an {\it end-to-end} manner. Pointing out the slow convergence and high computational cost of self-attention on image features, Zhu~\etal~\cite{zhu2020deformable} introduced multi-head deformable attention, replacing the dot-product in the attention computation with two linear projections for sampling points and computing their attention weights. While improving in both the convergence rate and accuracy, the strategy of sampling positions around a reference point prevents it to efficiently capture object information like object size and location. As sampled points on the image feature maps are separated, the module is unaware of the local connectivity of the attended region. Our \ours closely follows the overall framework of end-to-end object detection by Carion~\etal~\cite{nicolas2020detr}, but differs at its core by the use of the spatial prior and the multi-head box-attention mechanism. Our multi-head box-attention is inspired by the standard multi-head attention and convolution operation, which have both been shown to learn robust image representation. The box-attention considers a box region by only predicting its center and size, which is more efficient and allows us to extract structured information within the predicted region.

\boldparagraph{Attention for Instance Segmentation.} A method for tackling instance segmentation is required to locate objects and segment the pixels belonging to the object at the same time. Inspired by modern object detectors, earlier studies \cite{dai2016instanceconv,pinheiro2016refineobject} predict segment proposals in a first stage; the segment proposals are then classified in a second stage. He~\etal~\cite{he2017maskrcnn} proposed to train object detection and instance segmentation simultaneously in a multitask setting to boost the performance of both tasks. Different from modern segmentation models, which predict bounding boxes and masks from the same set of features (\ie, ResNet features), DETR relies on transformer features for object detection and ResNet features augmented with attention maps from the transformer for segmentation. This causes a mismatch in information level since these two tasks are highly related. Dong \etal \cite{dong2021solq} suggested to learn unified queries for both object detection and instance segmentation by taking advantage of deformable attention. However, this approach still lags behind convolution-based architectures by a large margin. We introduce box-attention which naturally extends to both object detection and instance segmentation in a single \ours-2D architecture achieving state-of-the-art performance on both tasks.

\boldparagraph{Attention for 3D Object Detection.} The main challenge in 3D object detection is to deal with rotated bounding boxes from bird's-eye view image features. Many methods \cite{sun2020waymodata,shi2020pvrcnn,li2020lidarrcnn} adapted Faster R-CNN by generating anchors of different angles as object proposals, followed by classification and regression. As anchor-based methods generate a high number of overlapping proposals, which requires non-maximum suppression to be tuned for each of the classes, approaches in \cite{ge2020afdet,yin2021center} focused on predicting a heat-map of object centers in a scene. While the number of overlapping proposals is reduced, predicting a heat-map leads to the loss of prior information compared to anchors and still relies on non-maximum suppression to filter out object proposals. A transformer with self-attention was also adopted for 3D object detection in \cite{liu2021grouptransformer,sheng2021channeltransformer}. Unfortunately, they exhibit the same problems as traditional detectors since their methods require initial object predictions from previous methods. The recent work of Misra~\etal~\cite{misra2021endtoend3d} introduced 3DETR for indoor 3D object detection. This method utilizes self-attention in both the encoder and decoder with object queries generated by a Farthest Point Sampling algorithm on point clouds \cite{qi2017farthestpoint}. 
Instead, \ours presents a solution for end-to-end 3D object detection on outdoor scenes that simply uses bird's-eye view features to predict objects without non-maximum suppression, 3D rotated IoU, or a complicated initialization method.

%% file: box_attention.tex
Box-attention is a multi-head attention mechanism designed to attend to boxes of interest in an image feature map. To do so, it samples a grid within each box and computes attention weights on sampled features from the grid structure, making the module easy to generalize to 2D or 3D object detection as well as instance segmentation. In each head of the attention computation, a box of interest is generated by predicting geometric transformations (\ie, translation, scaling, and rotation) from a pre-defined reference window.
The box-attention design allows the network to attend to dynamic regions of image features with reasonable computational cost.

\boldparagraph{Multi-Head Self-Attention.} We start by briefly summarizing the standard multi-head self-attention in the Transformer~\cite{vaswani2017transformer}. 
The multi-head self-attention of $l$ attention heads generates output features of the queries $(Q)$ by calculating weighted average vectors of the value features $(V)$ corresponding to the key vectors $(K)$:
\begin{equation}
\label{eqn:multi_head}
\mathop{\mathrm{MultiHead}}(Q, K, V) = \mathop{\mathrm{Concat}}({\it h}_1, \ldots, {\it h}_l) \, W^O \enspace,
\end{equation}
where ${\textit{h}}_i {=} \mathop{\mathrm{Attention}}(QW^Q_i, KW^K_i, VW^V_i)$. 
The self-attention module computes an attention map in each head using the dot-scale product of features between $Q$ and $K$, in which the computation increases quadratically with the matrix size.
\begin{equation}
\label{eqn:dot_scale}
\mathop{\mathrm{Attention}}(Q, K, V) = \softmax\left(\frac{QK^{\top}}{\sqrt{d_k}}\right) \, V \enspace,
\end{equation}
where $d_k$ is the dimension of the key feature in one head.

\boldparagraph{Multi-Head \boxattn.} Box-attention adopts the multi-head attention computation in Eq.~\eqref{eqn:multi_head} with the same feature aggregation of multiple heads and a learnable projection matrix $W^O$. 
In the attention computation stage, given a box of interest $b_i$ of query vector $q \in \mathbb{R}^{d}$ in the $i^{\text{th}}$ attention head, box-attention extracts a grid feature map $v_i$ of size $m {\times} m$ from $b_i$ using bilinear interpolation as illustrated in Fig.~\ref{fig:box_attn}.
The use of bilinear interpolation to compute the exact values of the grid features reduces the quantization error of the box-attention in box regression and pixel segmentation.
This differs from deformable attention~\cite{zhu2020deformable}, which predicts unstructured points causing ambiguity in capturing object information.
Instead, our attention mechanism inherits the spirit of RoIAlign~\cite{he2017maskrcnn} that precisely samples a grid structure within a region of interest (\ie, bounding box proposals) to obtain accurate pixel-level information which has been found to be important for pixel-accurate masks.

During the $i^{\text{th}}$ head attention computation, we treat the grid feature map $v_i \in \mathbb{R}^{m \times m \times d_h}$ as a set of value features corresponding to the query $q \in \mathbb{R}^{d}$.
The $m {\times} m$ attention scores are then generated by computing the dot-product between $q$ and $m {\times} m$ learnable key vectors $K_i$, where each vector represents a relative position in the grid structure, followed by a $\softmax$ function.
Thus, we share the same set of keys across queries.
By treating $K_i$ as relative location embedding of the sampled grid, box-attention can efficiently capture spatial information regarding the region.
In the implementation, the attention map generation can be performed efficiently via a simple linear projection ($\mathbb{R}^d \rightarrow \mathbb{R}^{m \times m}$) that is equivalent to the dot-product with learnable key vectors.
The final $h_{i} \in \mathbb{R}^{d_h}$ is the weighted average of the $m {\times} m$ vectors in $v_i$ with attention weights.
\begin{equation}
\begin{split}
\label{eqn:boxattn}
h_i & = \mathop{\mathrm{\boxattnop}}(Q, K_i, V_i) \\
    & = \sum_{m \times m} \softmax \big(QK_i^{\top}\big) * V_{i} \enspace,
\end{split}
\end{equation}
where $Q \in \mathbb{R}^{N \times d}$, $K_i \in \mathbb{R}^{(m \times m) \times d}$, $V_i \in \mathbb{R}^{N \times (m \times m) \times d_h}$, and $d_h$ is the dimension of features in one head.

It has been shown in \cite{tsung2017fpn} that multi-scale feature maps lead to large improvements in both object detection and instance segmentation. 
Our box-attention can be simply extended to work on multi-scale features. Given a set of boxes $\{b_i^1, \ldots, b_i^t\}$ of the query vector $q$ in an attention head, each of which belongs to each of $t$ multi-scale feature maps, we sample a grid of features from each box, resulting in $v_i \in \mathbb{R}^{(t \times m \times m) \times d_h}$. The $t {\times} m {\times} m$ attention scores are computed in the same way with $t {\times} m {\times} m$ learnable key vectors $K_i \in \mathbb{R}^{(t \times m \times m) \times d}$, where each vector represents a relative position in $t$ grid structures, followed by a $\softmax$ normalization. The $h_{(1, \ldots, l)} \in \mathbb{R}^{d_h}$ feature now is the weighted average of $t {\times} m {\times} m$ vectors in $v_{(1, \ldots, l)}$ as in Eq.~\eqref{eqn:boxattn}.

\boldparagraph{Multi-Head \instanceattn.} Instance-attention is a simple extension of box-attention without any extra parameters.
Our goal is to generate an accurate mask from the box of interest for instance segmentation.
In the $i^{\text{th}}$ attention head, it generates two outputs, $h_i \in \mathbb{R}^{d_h}$ for object detection and $h_i^\text{mask} \in \mathbb{R}^{m \times m \times {d_h}}$ for instance segmentation.
While weighted-averaging the $t {\times} m {\times} m$ features in $v_i$ to create $h_i$, we collapse $v_i$ in the first dimension (which contains the number of multi-scale features) for $h_i^\text{mask}$.
To do this, we normalize the first dimension of the $t {\times} m {\times} m$ attention scores using the $\softmax$ function which are then applied to $v_i$.
Note that we share all parameters of the attention module in generating $h_{(1,\ldots,l)}$ and $h^\text{mask}_{(1,\ldots,l)}$,  including the learnable projection matrix $W^O$.

\boldparagraph{Where-to-attend.}
Where-to-attend is a key component of our box-attention, it refers to an operation for predicting a box of interest in the attention computation. Specifically, the module learns to transform a reference window of query $q$ on a feature map into an attended region via simple geometric transformations, such as translation and scaling (see Fig. \ref{fig:box_attn}).

To be specific, we denote the reference window of query $q$ by $b_{q} {=} [x, y, w_x, w_y] \in [0, 1]^4$ where $x, y$ indicate its center position, $w_x, w_y$ are width and height of the window in normalized coordinates. The translation function, $\mathcal{F}_t$, takes $q$ and $b_{q}$ as its inputs and performs translation, which outputs $b'_{q}$ as follows:
\begin{equation}
\label{eqn:translation}
    \mathcal{F}_t(b_{q}, q) = b'_{q} = [x + \Delta_x, y + \Delta_y, w_x, w_y] \enspace,
\end{equation}
where $\Delta_x$ and $\Delta_y$ are offsets %of the center in the $x$ and $y$ axis in the coordinate of reference window.
relative to the center of the reference window.
Similarly, the scaling function, $\mathcal{F}_s$, takes the same inputs and adjusts the size of $b_{q}$
\begin{equation}
\label{eqn:scaling}
    \mathcal{F}_s(b_{q}, q) = b'_{q} = [x, y, w_x + \Delta_{w_x}, w_y + \Delta_{w_y}] \enspace,
\end{equation}
where $\Delta_{w_x}$ and $\Delta_{w_y}$ are offsets %of the size in the $x$ and $y$ axis in the coordinate of reference window.
for the reference window size.
The offset parameters (\ie, $\Delta_x, \Delta_y, \Delta_{w_x}, \Delta_{w_y}$) are predicted using a linear projection on $q$ for efficiency.
In the multi-head attention setting of $l$ heads and $t$ multi-scale feature maps, we use $l {\times} t$ transformation functions where each function predicts a box of interest $b_i^j$ for $i^{\text{th}}$ head and $j^{\text{th}}$ feature map.

% \jihong{As discussed, this $w_q$ in equation (5) and (6) is in fact the box, denoted as b previously. A window contains also the grids within the box. We could still use window in other places. But it would be better to make it explicit that the window is parametrized by the box.} (solved)

Where-to-attend is a combination of transformations and allows our box-attention to effectively attend to necessary regions with a small number of parameters and low computational overhead.
It can also be seen as a pseudo prediction step since it provides the network spatial information to predict a region of interest within the attention module.

%% file: boxer2d.tex
\begin{figure}[t]
\centering
\includegraphics[width=0.9\linewidth]{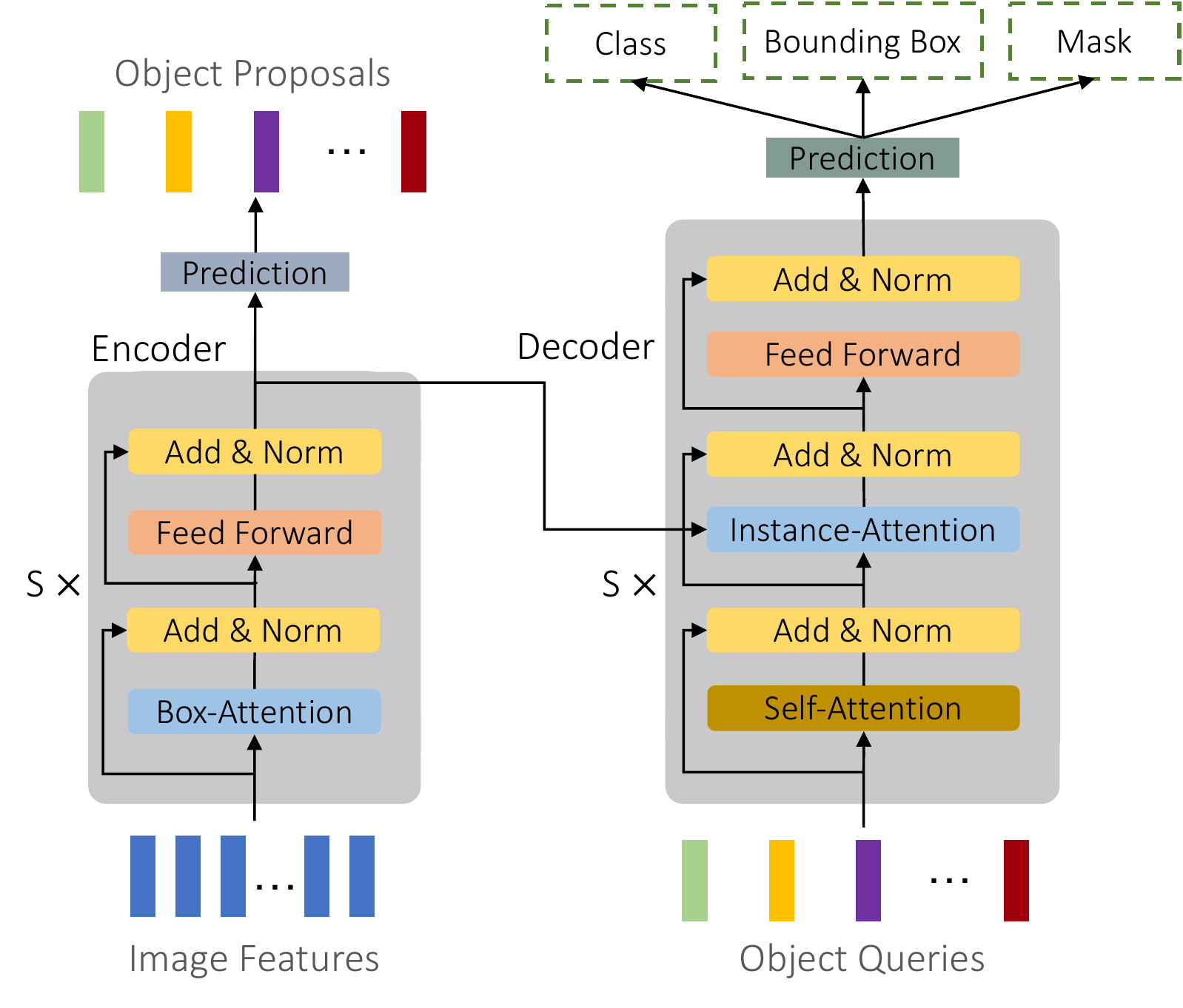}\\[-10pt]
\caption{
\textbf{\ours structure.} The \ours-2D takes encoder features corresponding to object proposals as its object queries. The object queries are then decoded into bounding boxes and pixel masks using instance-attention.}
\vspace{-0.2in}
\label{fig:boxer}
\end{figure}

\begin{figure*}[t]
\centering
\includegraphics[width=1\linewidth]{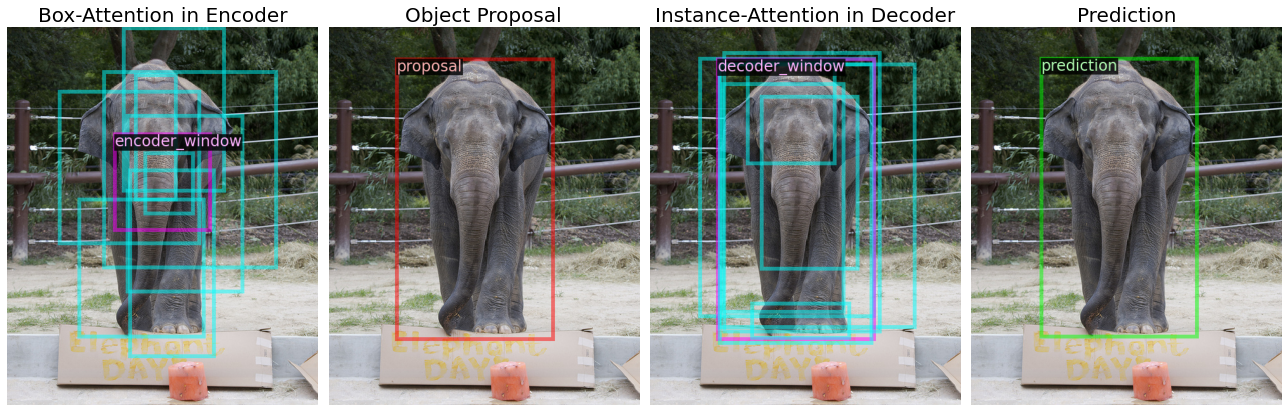}\\[-8pt]
\caption{\textbf{\ours-2D behavior.} We show the behavior of \ours-2D by tracing back its prediction. \boxattn in the encoder is able to capture regions of multiple aspect ratios from the reference window (denoted in purple), while \instanceattn in the decoder plays a role to refine the object proposal. The \ours-2D predicts an object proposal which highly overlaps with the final prediction.}
\label{fig:vis_boxer}
\vspace{-0.2in}
\end{figure*}

To demonstrate the effectiveness of our approach, we present \ours, a Transformer-based network with box-attention in its architecture; see Fig.~\ref{fig:boxer}.
\ours consists of an encoder for encoding multi-scale feature maps extracted from a backbone and a decoder for predicting instance bounding boxes and pixel masks.
Our approach follows the spirit of end-to-end object detection frameworks (\ie, DETR~\cite{nicolas2020detr}), that reduce the need for hand-crafted modules, like non-maximum suppression and anchor-groundtruth matching.

\boldparagraph{\ours Encoder.} As in Transformer, each \ours encoder layer contains box-attention and feed forward sub-layers, each of which is followed by a LayerNorm~\cite{ba2016layernorm} with residual connection.
Following \cite{zhu2020deformable}, the encoder takes multi-scale image feature maps $\{x^j\}^{t-1}_{j{=}1} (t{=}4)$ extracted from $C_3$ through $C_5$ of a ResNet backbone \cite{kaiming2016resnet} (transformed by a $1 {\times} 1$ convolution to the hidden dimension) as its inputs. 
The $t^\text{th}$ feature map $x^t$ is obtained by applying a $3 {\times} 3$ convolution layer with stride 2 on the final $C_5$ feature. 
The \ours encoder will transform multi-scale inputs into multi-scale contextual representations $\{e^j\}^t_{j{=}1}$.
Note that the multi-scale contextual representations $\{e^j\}^t_{j{=}1}$ are in the same resolution as the inputs $\{x^j\}^{t}_{j{=}1}$.

In the encoder, both $Q$ and $V$ are features from multi-scale feature maps.
We assign a reference window to each query vector %where the center of the window is the pixel position.
where the window is centered at the query spatial position.
The sizes of the sliding windows are $\{32^2, 64^2, 128^2, 256^2\}$ pixels on multi-scale feature maps $\{x_1, x_2, x_3, x_4\} (t{=}4)$ (or $4^2$ features on each of the multi-scale feature maps), as suggested in \cite{tsung2017fpn}.
Because $l$ parallel attention heads of box-attention are able to implicitly capture boxes of multiple aspect ratios at each feature level, we found that it is not necessary to have reference windows of multiple aspect ratios (see Fig. \ref{fig:vis_boxer}).
Beside augmenting each query with a position embedding, we add a size embedding, which represents the size of the reference window corresponding to each query.
The size embedding only differs between query vectors of different levels.
Both embeddings are normalized and encoded with sinusoid encodings.

Since two-stage networks indicate a significant improvement in object detection \cite{ren2015faster_rcnn,zhu2020deformable}, we show that the \ours encoder is able to produce high-quality object proposals as inputs for the decoder.
In the object proposal stage, features from the last encoder layer are fed into a prediction head to predict object proposals \wrt their reference windows.
Instead of treating the sinusoid embedding of bounding boxes predicted in the object proposal stage as object queries \cite{zhu2020deformable}, we simply take the encoder features (transformed by a linear projection) with the highest classification scores as input features for the decoder.
This provides richer information to the \ours decoder as encoder features contain both spatial and contextual information.
The predicted bounding boxes are treated as reference windows for its corresponding proposals in the decoder.

\boldparagraph{\ours Decoder.} In each \ours decoder layer, the cross-attention sub-layer is our multi-head instance-attention, while the self-attention and feed forward sub-layers are left unchanged.
The features of the object proposals from the encoder are the inputs of \ours decoder.
The reference windows of the object proposals are refined in this stage in order to give accurate predictions.

To be specific, we denote the inputs to the $(s + 1)^{\text{th}}$ decoder layer by $x_s \in \mathbb{R}^{N \times d}$.
The $(s + 1)^{\text{th}}$ decoder layer then outputs $x_{s+1} \in \mathbb{R}^{N \times d}$ and $x^\text{mask}_{s+1} \in \mathbb{R}^{N \times m \times m \times d}$.
The feed forward sub-layer is the same for both outputs.
The output features $x_S \in \mathbb{R}^{N \times d}$ are then decoded into box coordinates and class labels as in \cite{zhu2020deformable}, while $x^\text{mask}_{S} \in \mathbb{R}^{N \times m \times m \times d}$ are used to generate instance masks.
We follow the training strategy in Mask R-CNN~\cite{he2017maskrcnn} to predict instance masks with a per-pixel {\it sigmoid} and a {\it binary} loss. 

Since the where-to-attend module in the attention module predicts regions of interest based on reference windows, we design the detection head to predict a bounding box as a relative offset \wrt its reference window size and position.
The reference window serves as an initial guess of its object proposal feature in the prediction stage.
The auxiliary decoding losses for other decoder layers are also effective in our case.
All prediction heads in the \ours decoder share their parameters.
We found that it is not necessary to add a mask cost into the Hungarian matcher~\cite{kuhn1955hungarian}, which results in a more efficient training.
More details are provided in the supplementary document.

%% file: boxer3d.tex
We enable end-to-end 3D object detection by extending our box-attention and \ours to work with point cloud input.

\boldparagraph{\boxattn for 3D Object Detection.} Along with translation and scaling in the where-to-attend module, we add rotation transformation in the bird's-eye view plane to model the angle of objects.
We denote the reference window of $q$ by $b_{q} {=} [x, y, w_x, w_y, \theta] \in [0, 1]^5$ where $x, y$ indicate its center position, $w_x, w_y$ are width and height of the window, and $\theta$ is the rotation angle of $b_{q}$ around its center in normalized coordinates.
The final rotation function, $\mathcal{F}_r$, predicts an offset of the window rotation angle. It then applies a rotation matrix on the $m {\times} m$ grid coordinates sampled from $b_{q}$
\begin{equation}
\label{eqn:rotation}
    \mathcal{F}_r(b_{q}, q) = b'_{q} = [x, y, w_x, w_y, \theta + \Delta_{\theta}] \enspace,
\end{equation}
where $\Delta_{\theta}$ is an offset \wrt the reference window angle. Again, we use a linear projection on $q$ to predict $\Delta_{\theta}$.

\begin{figure}[t]
\centering
\includegraphics[width=1\linewidth]{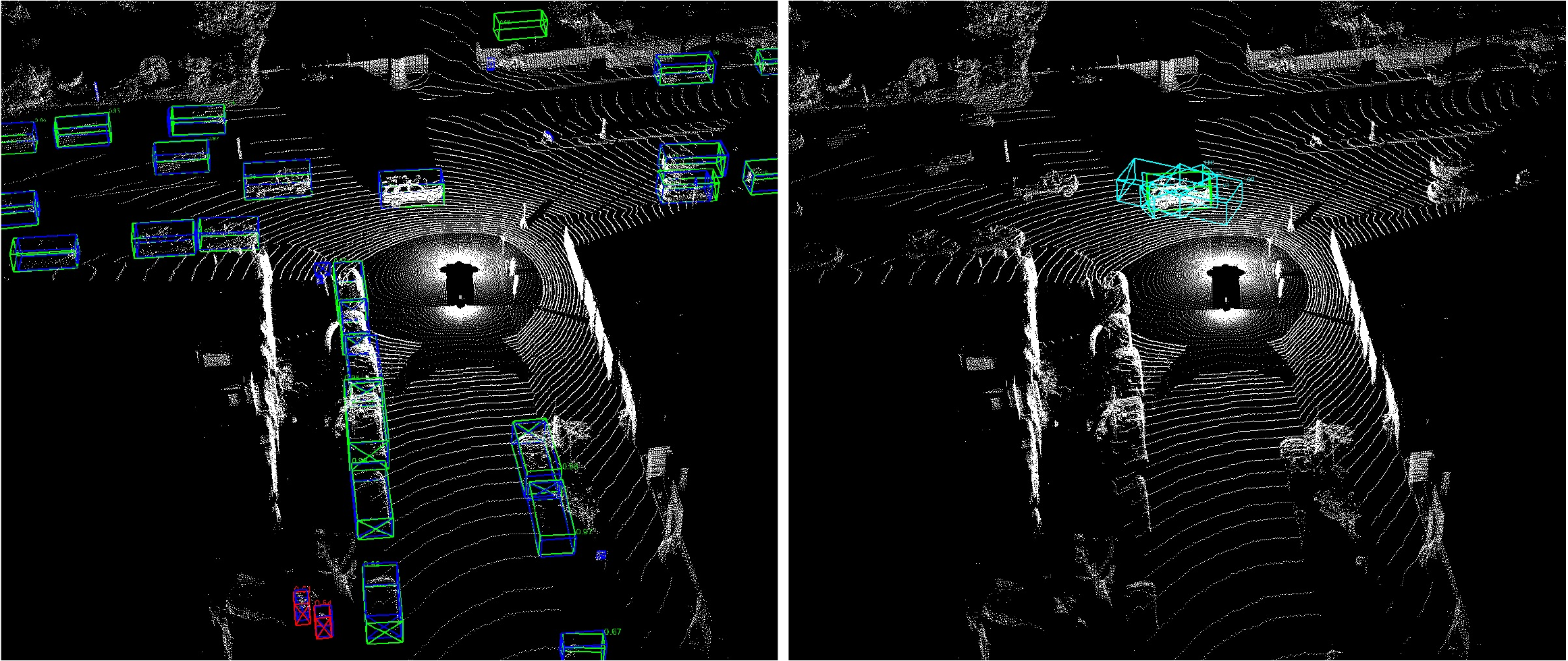}\\[-8pt]
\caption{\textbf{\ours-3D prediction.} Left: \ours-3D prediction in an intersection (ground-truth boxes denoted in blue; vehicle and pedestrian predictions denoted in green and red). Right: Visualization of \boxattn behavior corresponding to one object query. Multiple attention heads of the object query in \boxattn capture boxes of different angles where the best attended region is well-aligned with \ours-3D prediction.}
\label{fig:vis_3d}
\vspace{-0.2in}
\end{figure}

\boldparagraph{\ours for 3D Object Detection.} To better capture objects of different angles, we assign reference windows of multiple angles to each query vector of \ours encoder features.
At each sliding position, based on the 2D object detection setting, we use three reference windows of $4^2$ features on each of the multi-scale feature maps with three angles $\{\frac{-2\pi}{3}, 0, \frac{2\pi}{3}\}$. Each attention head will be assigned a reference window of one angle. By doing so, features generated from our box-attention are strong for rotation prediction (see Fig. \ref{fig:vis_3d}). In the object proposal stage, for each of the encoder features, we predict class scores and bounding boxes for the three proposals \wrt their reference windows of three angles. The 3D Hungarian matcher is used during training. More details are provided in the supplementary document.

We note that only {\it minimal} prior knowledge about specific object classes, such as the typical size of a vehicle is embedded in our system due to the uniform distribution of the reference window. This is different from previous methods \cite{sun2020waymodata,sun2021rangesparsenet,shi2020pvrcnn,yin2021center} which use different anchor sizes, heat-maps, or backbones for each class. Our network also removes the need for hand crafted modules such as rotated non-maximum suppression or 3D IoU computation.

%% file: experiments.tex
\subsection{Datasets, Tasks  and Implementation Details}

% -------------------------------------------
\boldparagraph{MS-COCO 2017.} For 2D object detection and instance segmentation, we use the MS-COCO 2017 dataset \cite{lin2014mscoco} consisting of 118,000 training images and 5,000 validation images. 
% Each image contains 7 instances on average, with a maximum number of 63 instances in the training split. 
The instance is categorized based on its size: small, medium and large. 
We report the standard COCO metrics for bounding boxes and masks. We use the \train split for training and report ablations on the \val split. We also report results on the \test set.

We use the Adam optimizer~\cite{kingma2015adam} with $\alpha{=}0.0002$, and weight decay set to $0.0001$. The learning rate of our backbone and transformation functions in the attention module is multiplied by a factor of 0.1. We find that dropout is not necessary for \ours and makes the training slower. Thus, we remove it from the \ours architecture. We train our network for 50 epochs with a batch size of 32, the learning rate is decayed at the 40$^\text{th}$ epoch by a factor of 0.1. Other hyper-parameter settings follow Deformable DETR~\cite{zhu2020deformable}. During the training procedure, the same data augmentation is used as in \cite{nicolas2020detr}. For a better comparison, we also report \ours-2D trained with a 3$\times$ schedule as in \cite{wu2019detectron2}.

%%%%%%%%%%%%%%%%%%%%%%%%%%%%%%%%%%%%%%%%%%%%%%%%%%%%%%%%%%%%%%%%%%%%
\begin{table}[!b]
\centering
{
\resizebox{1.0\linewidth}{!}{
\begin{threeparttable}
\tablestyle{5pt}{1.1}
\begin{tabular}{l|c|c|c|c|c}
  & FLOPs$\downarrow$ & AP$\uparrow$ & $\text{AP}_{\text{S}}\!\uparrow$ & $\text{AP}_{\text{M}}\!\uparrow$ & $\text{AP}_{\text{L}}\!\uparrow$\\
% \hshline
% (dynamic)  & \cmark & \xmark & \cmark & \xmark & \cmark & \xmark & \cmark \\
\shline
Self-Attention \cite{vaswani2017transformer} & 187G & 36.2 & 16.3 & 39.2 & 53.9 \\
Deformable-Attention$^{\dagger}$ \cite{zhu2020deformable} & 173G & 46.9 & 29.6 & 50.1 & 61.6 \\
Dynamic-Attention \cite{dai2021dynamidetr} & - & 47.2 & 28.6 & 49.3 & 59.1 \\
\hline
{\bf Box-Attention (Ours)} & 167G & \bf 48.7 & \bf 31.6 & \bf 52.3 & \bf 63.2  \\ % 25G more with mask prediction in new implementation
% w/o ($\mathcal{F}_t$ and $\mathcal{F}_s$) & 166G & 45.6 & 28.7 & 48.7 & 59.7  \\
w/o ($\mathcal{F}_t$ and $\mathcal{F}_s$) & 164G & 46.4 & 29.6 & 49.8 & 59.7 \\
\end{tabular}
\footnotesize{$^{\dagger}$~Based on author-provided github, which is higher than in their original paper. }
\end{threeparttable}
}
}
%\medskip
\vspace{-0.1in}
\caption{\textbf{\boxattn vs. alternatives} in end-to-end object detection on the COCO \val set using a R-50 backbone pretrained on ImageNet. Box-Attention performs best with the least FLOPs.}
\label{tab:boxattn}
% \vspace{-0.2in}
\end{table}
%%%%%%%%%%%%%%%%%%%%%%%%%%%%%%%%%%%%%%%%%%%%%%%%%%%%%%%%%%%%%%%%%%%%

%%%%%%%%%%%%%%%%%%%%%%%%%%%%%%%%%%%%%%%%%%%%%%%%%%%%%%%%%%%%%%%%%%%%
\begin{table*}[t]
\centering
{
% \resizebox{\linewidth}{!}{

\subfloat[\label{tab:proposals} {\bf Object Proposals.} ]
{\scalebox{0.78}{
\tablestyle{6pt}{1.0}
\begin{tabular}{l|c|c|c|c}
  & AP$\uparrow$ & $\text{AP}_{\text{S}}\!\!\uparrow$ & $\text{AP}_{\text{M}}\!\!\uparrow$ & $\text{AP}_{\text{L}}\!\!\uparrow$ \\
% \hshline
% (dynamic)  & \cmark & \xmark & \cmark & \xmark & \cmark & \xmark & \cmark \\
\shline
\boxattn & \bf 48.7 & \bf 31.6 & \bf 52.3 & \bf 63.2 \\
% + {\it dropout} & 47.9 & 31.4 & 50.9 & 62.5 \\
w/ {\it proposal refinement}  & 47.2 & 30.4 & 50.7 & 62.2 \\
% w/o {\it proposal refinement} & 47.9 & 31.4 & 50.9 & 62.5  \\
\end{tabular}
}}
\hspace{0.1in}%
\subfloat[\label{tab:multitask} {\bf \instanceattn.}]
{\scalebox{0.745}{
\tablestyle{6pt}{1.1}
\begin{tabular}{l|c|c|c|c|c|c|c|c}
 & AP$\uparrow$ & $\text{AP}_{\text{S}}\!\!\uparrow$ & $\text{AP}_{\text{M}}\!\!\uparrow$ & $\text{AP}_{\text{L}}\!\!\uparrow$ & AP$^\text{m}\!\!\uparrow$ & $\text{AP}^\text{m}_{\text{S}}\!\!\uparrow$ & $\text{AP}^\text{m}_{\text{M}}\!\!\uparrow$ & $\text{AP}^\text{m}_{\text{L}}\!\!\uparrow$ \\
\shline
\boxattn {\it \ only} & 48.7 & 31.6 & 52.3 & 63.2 & - & - & - & - \\
w/ \instanceattn & \bf 50.0 & \bf 32.4 & \bf 53.3 & \bf 64.5 & \bf 42.7 & \bf 22.7 & \bf 45.9 & \bf 61.5 \\
\end{tabular}
}}
}
% }
\hspace{0.1in}
%
%\medskip
\vspace{-0.8em}
\caption{\label{tab:boxer}\textbf{\ours-2D ablation} on the COCO \val set using a R-50 backbone pretrained on ImageNet. (a) Our reference windows improve the quality of the object proposals and removes the need for the refinement stage of \cite{zhu2020deformable}. (b) \ours-2D shows strong results when training on both 2D object detection and instance segmentation simultaneously.}
\vspace{-0.2in}
\end{table*}
%%%%%%%%%%%%%%%%%%%%%%%%%%%%%%%%%%%%%%%%%%%%%%%%%%%%%%%%%%%%%%%%%%%%

\boldparagraph{Waymo Open.} For 3D object detection, we use the Waymo Open dataset~\cite{sun2020waymodata}, which contains 798 training sequences and 202 validation sequences. Each sequence consists of 200 frames where each frame captures the full 360 degrees around a vehicle. 
We report the official 3D detection evaluation metrics including the standard 3D bounding box mean average precision (mAP) and mAP weighted by heading accuracy (mAPH) in three categories: vehicle, pedestrian, and cyclist.
% There are two difficulty levels: LEVEL\_1 for boxes with more than five Lidar points and LEVEL\_2 for boxes with at least one Lidar point.

We use the Adam optimizer with weight decay set to 0.0001. Following previous works \cite{sun2021rangesparsenet}, we use cosine learning rate decay with the initial learning rate set to 5e-4, 5000 warm-up steps starting at 5e-7, and 140K iterations in total. The learning rate of the transformation functions in the attention module is multiplied by a factor of 0.1. We train our network on BEV image features extracted from PointPillar~\cite{lang2019pointpillar} with a grid size of $(0.32\text{m}, 0.32\text{m})$. The detection range is $[-75.0\text{m}, 75.0\text{m}]$ for the $x$ and $y$ axis, and $[-4\text{m}, 8\text{m}]$ for the $z$ axis. For ablation studies, we train our network on only 20\% of the training data.

%CS: already in abstract.
%\cs{Add statement on availability of code?}

% -------------------------------------------
\subsection{Ablation Study}

%We first evaluate the contribution of the \ours-2D and \ours-3D network design. 
%For 2D object detection and segmentation, we conduct our experiments using an R-50 backbone pretrained on ImageNet. 
%The ablation studies of 2D object detection and instance segmentation will be reported on the COCO \textit{val} set while the 3D object detection results will be reported on the Waymo \textit{val} set.

\boldparagraph{\boxattn vs. Alternatives.}
We first compare \boxattn with Self-Attention \cite{vaswani2017transformer}, Deformable-Attention \cite{zhu2020deformable} and Dynamic-Attention \cite{dai2021dynamidetr} in end-to-end object detection. Results in Table \ref{tab:boxattn} indicate an improvement for \boxattn on all metrics, with the highest gain from \textit{small} objects (AP$_\text{S}$) (up to 2 points). Furthermore, the \boxattn requires a smaller number of FLOPs compared to other attention mechanisms. We also report \boxattn \textit{without} the where-to-attend module that adopts the reference window but not the transformation functions (translation and scaling). It can be seen in Table \ref{tab:boxattn} that the where-to-attend module contributes more than 2 points in all categories. This shows the importance of translation and scaling functions in learning to attend to the relevant region.

%%%%%%%%%%%%%%%%%%%%%%%%%%%%%%%%%%%%%%%%%%%%%%%%%%%%%%%%%%%%%%%%%%%%
\begin{table}[b!]
\centering
\footnotesize
%{
%\resizebox{0.9\linewidth}{!}{
% \subfloat[\label{tab:rotation} {\bf Where-to-attend with rotation.}]
% {
\tablestyle{5.4pt}{1.1}
\begin{tabular}{l|cc|cc|cc}
  & \multicolumn{2}{c|}{Vehicle} & \multicolumn{2}{c|}{Pedestrian}& \multicolumn{2}{c}{Cyclist}  \\
  & AP$\uparrow$ & APH$\uparrow$ & AP$\uparrow$ & APH$\uparrow$ & AP$\uparrow$ & APH$\uparrow$ \\
\shline
$\mathcal{F}_r$ + multi-angle & \bf 70.4 & \bf 70.0 & \bf 64.7 & 53.5 & \bf 50.2 & \bf 48.9 \\
\hline
w/o $\mathcal{F}_r$    & 69.4 & 68.7 & 63.3 & 52.8 & 47.4 & 46.1 \\
w/o multi-angle        & 70.0 & 69.3 & \bf 64.7 & \bf 53.7 & 48.2 & 47.0 \\
\end{tabular}
% }
%}
%
%}
%\medskip
\vspace{-8pt}
\caption{\textbf{\ours-3D ablation} on the Waymo \val set (LEVEL\_1 difficulty). Adding $\mathcal{F}_r$ gives better performance for detecting 3D bounding boxes. Multi-angle reference windows further improve results by taking advantage of an explicit angle prior.}
\label{tab:boxer3d}
\vspace{-0.2in}
\end{table}
%%%%%%%%%%%%%%%%%%%%%%%%%%%%%%%%%%%%%%%%%%%%%%%%%%%%%%%%%%%%%%%%%%%%

\boldparagraph{\ours-2D Ablation.} 
%Table \ref{tab:boxer} shows ablations of \ours-2D with various design choices for 2D object detection and instance segmentation. 
As \ours-2D utilizes multi-scale reference windows in its encoder for predicting object proposals, these  proposals serve as reference windows in the decoder. In Table~\ref{tab:proposals}, we evaluate the quality of our object proposals by adding object proposal refinement in the decoder layers. While such refinement proved beneficial in \cite{zhu2020deformable}, we observe more than 1 point drop in AP. This suggests that when object proposals are generated by the \ours-2D encoder with reference windows, they are sufficient for the \ours-2D decoder to predict objects without the need for a refinement in each step (see Fig. \ref{fig:vis_boxer}). Our \ours-2D is flexible, as we can easily plug \instanceattn into its decoder in order to predict both the object location and its overlay. Table~\ref{tab:multitask} shows \ours-2D benefits from multi-task training (object detection and instance segmentation). Note that this is not the case for DETR~\cite{nicolas2020detr}. In our setting, the multi-task training does not require more parameters except for a small mask prediction head. The training is also stable without any change in hyper-parameters.

\boldparagraph{\ours-3D Ablation.}
We ablate the effectiveness of our \ours-3D design on 3D object detection in Table~\ref{tab:boxer3d}. 
The table indicates the role of rotation transformation in the \textit{where-to-attend} module, which contributes more than 1 point in all categories at the expense of a small amount of computation. Specifically, we found rotation transformation is most effective when added to box-attention in the decoder layers.
Table~\ref{tab:boxer3d} also shows the comparison between multi-angle \vs single-angle reference window in the \ours-3D encoder layers. Using a multi-angle reference window yields an improvement for the vehicle and cyclist category, while remaining stable for pedestrians. This suggests that each head in multi-head attention is able to effectively capture the information of different rotation angles.

% -------------------------------------------
\subsection{Comparison with Existing Methods}

%%%%%%%%%%%%%%%%%%%%%%%%%%%%%%%%%%%%%%%%%%%%%%%%%%%%%%%%%%%%%%%%%%%%
% \begin{table*}
% \centering
% {
% \resizebox{0.75\linewidth}{!}{
% \subfloat[\label{tab:rotation} {\bf Where-to-attend with rotation.}]
% {
% \tablestyle{2.3pt}{1.2}
% \begin{tabular}{l|cc|cc|cc}
%  & \multicolumn{2}{c|}{Vehicle} & \multicolumn{2}{c|}{Pedestrian}& \multicolumn{2}{c}{Cyclist}  \\
%  & AP & APH & AP & APH & AP & APH \\
% % \hshline
% % (dynamic)  & \cmark & \xmark & \cmark & \xmark & \cmark & \xmark & \cmark \\
% \shline
% w/o $\mathcal{F}_r$ & 69.4 & 68.7 & 63.3 & 52.8 & 47.4 & 46.1 \\
% w/ $\mathcal{F}_r$  & 70.4 & 70.0 & 64.7 & 53.5 & 50.2 & 48.9 \\
% \end{tabular}
% }
% \hspace{0.2in}%
% \subfloat[\label{tab:multi_angle} {\bf Multi-angle reference window.} ]
% {
% \tablestyle{2.3pt}{1.2}
% \begin{tabular}{l|cc|cc|cc}
%  & \multicolumn{2}{c|}{Vehicle} & \multicolumn{2}{c|}{Pedestrian}& \multicolumn{2}{c}{Cyclist}  \\
%  & AP & APH & AP & APH & AP & APH \\
% \shline
% {\it single-angle} & 70.0 & 69.3 & 64.7 & 53.7 & 48.2 & 47.0\\
% {\it multi-angle}  & 70.4 & 70.0 & 64.7 & 53.5 & 50.2 & 48.9 \\
% \end{tabular}
% }
% }
% %
% }
% %\medskip
% \caption{\label{tab:boxer3d}\textbf{\ours-3D ablation.} (a) Adding rotation transformation gives better performance for detecting 3D bounding boxes. (b) Multi-angle reference windows improve 3D detection results as they take advantage of an explicit angle prior.}
% \vspace{-0.2in}
% \end{table*}
%%%%%%%%%%%%%%%%%%%%%%%%%%%%%%%%%%%%%%%%%%%%%%%%%%%%%%%%%%%%%%%%%%%%

%%%%%%%%%%%%%%%%%%%%%%%%%%%%%%%%%%%%%%%%%%%%%%%%%%%%%%%%%%%%%%%%%%%%
\begin{figure}[b]
\centering

\begin{minipage}{1\linewidth}
\centering

\begin{minipage}[b]{.49\linewidth}
\includegraphics[width=\linewidth]{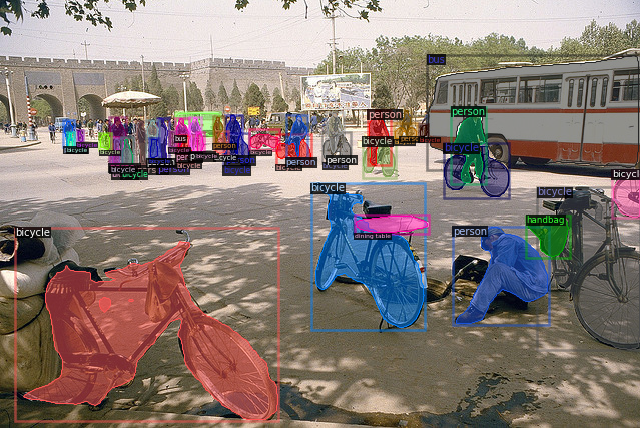}
\end{minipage}
\begin{minipage}[b]{.49\linewidth}
\includegraphics[width=\linewidth]{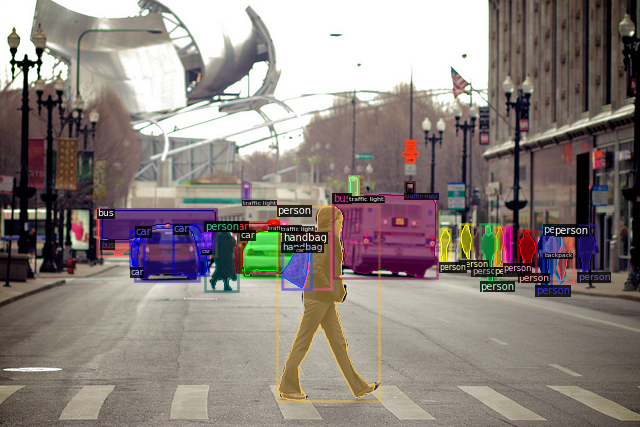}
\end{minipage}

\end{minipage}

\vspace{-0.6em}
\caption{\textbf{Qualitative results} for instance detection and segmentation in the COCO 2017 \test set generated by \ours-2D (More qualitative results are in the supplementary document).}
\label{fig:vis_2d}
% \vspace{-0.2in}
\end{figure}
%%%%%%%%%%%%%%%%%%%%%%%%%%%%%%%%%%%%%%%%%%%%%%%%%%%%%%%%%%%%%%%%%%%%

%%%%%%%%%%%%%%%%%%%%%%%%%%%%%%%%%%%%%%%%%%%%%%%%%%%%%%%%%%%%%%%%%%%
\begin{table*}
\centering
% \scriptsize
\footnotesize
{
\tablestyle{8pt}{1.2}
\begin{tabular}{l|c|c|c|cccccc}
Method & Backbone & Epochs & end-to-end & AP$\uparrow$ & $\text{AP}_{\text{50}}\!\!\uparrow$ & $\text{AP}_{\text{75}}\!\!\uparrow$ & $\text{AP}_{\text{S}}\!\!\uparrow$ & $\text{AP}_{\text{M}}\!\!\uparrow$ & $\text{AP}_{\text{L}}\!\!\uparrow$ \\
\shline
Faster RCNN-FPN \cite{ren2015faster_rcnn}  & R-101  & 36 & \xmark & 36.2 & 59.1 & 39.0 & 18.2 & 39.0 & 48.2 \\
ATSS \cite{zhang2020atss}   & R-101 &    24 & \xmark &       43.6 & 62.1 & 47.4 & 26.1 & 47.0 & 53.6 \\
Sparse RCNN \cite{sun2021sparse_rcnn} & X-101 & 36 & \cmark & 46.9 & 66.3 & 51.2 & 28.6 & 49.2 & 58.7 \\
VFNet \cite{zhang2020vfnet} & R-101 &  24 & \xmark &    46.7 & 64.9 & 50.8 & 28.4 & 50.2 & 57.6 \\ 
\hline
Deformable DETR \cite{zhu2020deformable} & R-50    & 50 & \cmark & 46.9 & 66.4 & 50.8 & 27.7 & 49.7 & 59.9 \\
Deformable DETR \cite{zhu2020deformable} & R-101    & 50 & \cmark  & 48.7 & 68.1 & 52.9 & 29.1 & 51.5 & 62.0 \\
Dynamic DETR \cite{dai2021dynamidetr} & R-50 & 50 & \cmark & 47.2 & 65.9 & 51.1 & 28.6 & 49.3 & 59.1 \\
TSP-RCNN \cite{sun2021rethinking}       & R-101 & 96 & \cmark &   46.6 & 66.2 & 51.3 & 28.4 & 49.0 & 58.5 \\
\hline
% \textbf{\ours-2D}                      & R-50  & 50  & 49.8 & 68.0 & 54.0 & 30.4 & 52.2 & 62.4 \\
\textbf{\ours-2D}                      & R-50  & 50 & \cmark  & 50.0 & 67.9 & 54.7 & 30.9 & 52.8 & 62.6 \\
\textbf{\ours-2D} (3$\times$ schedule) & R-50  & 36 & \cmark  & 49.9 & 68.0 & 54.4 & 30.9 & 52.6 & 62.5 \\
\textbf{\ours-2D} (3$\times$ schedule) & R-101 & 36 & \cmark  & \bf 51.1 & \bf 68.5 & \bf 55.8 & \bf 31.5 & \bf 54.1 & \bf 64.6 \\

\end{tabular}
}
%\medskip
\vspace{-0.8em}
\caption{\textbf{Comparison of \ours-2D in object detection} on the COCO 2017 \test set with various backbone networks. \ours-2D outperforms other methods including transformer-based object detectors with a faster training schedule.}
\label{tab:comp_det}
\vspace{-0.1in}
\end{table*}
%%%%%%%%%%%%%%%%%%%%%%%%%%%%%%%%%%%%%%%%%%%%%%%%%%%%%%%%%%%%%%%%%%%%

%%%%%%%%%%%%%%%%%%%%%%%%%%%%%%%%%%%%%%%%%%%%%%%%%%%%%%%%%%%%%%%%%%%
\begin{table*}
\centering
\footnotesize
{
\begin{threeparttable}
\tablestyle{6.8pt}{1.2}
\begin{tabular}{l|c|c|cccc|cccc}
 & Epoch & end-to-end & AP$\uparrow$ &  $\text{AP}_{\text{S}}\!\!\uparrow$ & $\text{AP}_{\text{M}}\!\!\uparrow$ & $\text{AP}_{\text{L}}\!\!\uparrow$ & AP$^\text{m}\!\!\uparrow$ & $\text{AP}^\text{m}_{\text{S}}\!\!\uparrow$ & $\text{AP}^\text{m}_{\text{M}}\!\!\uparrow$ & $\text{AP}^\text{m}_{\text{L}}\!\!\uparrow$ \\
\shline
Mask R-CNN \cite{he2017maskrcnn} & 36 & \xmark & 43.1 & 25.1 & 46.0 & 54.3 & 38.8 & 21.8 & 41.4 & 50.5 \\
% SOLOv2 \cite{} & 72 & 42.6 & 22.3 & 46.7 & 56.3 & 39.7 & 17.3 & 42.9 & 57.4 \\
QueryInst \cite{fang2021queryinst} & 36 & \xmark & 48.1 & - & - & - & 42.8 & 24.6 & 45.0 & 55.5 \\
\hline
SOLQ \cite{dong2021solq} & 50 & \cmark & 48.7 & 28.6 & 51.7 & 63.1 & 40.9 & 22.5 & 43.8 & 54.6 \\
\hline
\textbf{\ours-2D} (3$\times$ schedule) & 36 & \cmark & \bf 51.1 & \bf 31.5 & \bf 54.1 & \bf 64.6 & \bf 43.8 & \bf 25.0 & \bf 46.5 & \bf 57.9 \\
% \ours  & 49.0 & 30.5 & 52.3 & 63.6 & 38.6 & 19.3 & 41.7 & 57.1 \\
\end{tabular}
\end{threeparttable}
}
%\medskip
\vspace{-0.8em}
\caption{\textbf{Comparison of \ours-2D in instance segmentation} on the COCO 2017 \test set using a R-101 backbone. \ours-2D shows better results in both detection and instance segmentation.}
\label{tab:instance_segment}
\vspace{-0.2in}
\end{table*}
%%%%%%%%%%%%%%%%%%%%%%%%%%%%%%%%%%%%%%%%%%%%%%%%%%%%%%%%%%%%%%%%%%%%

\boldparagraph{2D Object Detection.} Table \ref{tab:comp_det} lists the performance of previous methods and \ours-2D using ResNet-50 and ResNet-101 backbones. The first part contains convolution-based object detectors while the second part focuses on transformer-based methods. Across backbones \ours-2D achieves better results on all metrics. Notably, \ours-2D outperforms other methods in detecting small objects, with more than 2 $\text{AP}_\text{S}$ points improvement compared to Deformable DETR. In addition, our network is able to converge quickly with the standard 3$\times$ schedule setting \cite{wu2019detectron2}. It is further worth to point out that \ours-2D trained with the 3$\times$ schedule reaches competitive results.

\boldparagraph{2D Instance Segmentation.} We compare \ours-2D with other instance segmentation methods. In Table \ref{tab:instance_segment}, the 3$\times$ schedule is used in the training of our network. \ours-2D improves on all of the metrics for bounding boxes and instance masks against QueryInst \cite{fang2021queryinst}. Furthermore, our method outperforms SOLQ \cite{dong2021solq}, a transformer-based method, by around 2 points on all categories. The visualization of the \ours-2D prediction can be seen in Fig. \ref{fig:vis_2d}.

\boldparagraph{3D Object Detection.} Table \ref{tab:3d_obj} shows the performance of \ours-3D and other 3D object detectors along with a naive implementation of Deformable DETR \cite{zhu2020deformable} as our baseline. It can be seen that \ours-3D consistently improves over the baseline on all metrics, specially for small objects like pedestrians. Our network reaches a competitive result compared to highly optimized methods in the vehicle category. However, there is still a gap between \ours-3D and previous methods in the pedestrian category. It should be noted that compared to others we only use \textit{minimal} prior knowledge per category.

%%%%%%%%%%%%%%%%%%%%%%%%%%%%%%%%%%%%%%%%%%%%%%%%%%%%%%%%%%%%%%%%%%%
\begin{table}
\centering
\footnotesize
{
% \resizebox{1.0\linewidth}{!}{
\tablestyle{4.5pt}{1.2}
\label{tab:compare}
\begin{tabular}{l|c|cc|cc}
 & \multirow{2}{*}{end-to-end} & \multicolumn{2}{c|}{Vehicle} & \multicolumn{2}{c}{Pedestrian}  \\
%  & \multicolumn{2}{c|}{L1} & \multicolumn{2}{c|}{L2} & \multicolumn{2}{c|}{L1} & \multicolumn{2}{c}{L2}  \\
 & & AP$\uparrow$ & APH$\uparrow$ & AP$\uparrow$ & APH$\uparrow$ \\
% \hshline
% (dynamic)  & \cmark & \xmark & \cmark & \xmark & \cmark & \xmark & \cmark \\
\shline
PointPillar~\cite{lang2019pointpillar} & \xmark  & 55.2 & 54.7 & 60.0 & 49.1  \\
PV-RCNN~\cite{shi2020pvrcnn}           & \xmark & \bf 65.4 & \bf 64.8 & - & -  \\
RSN S\_1f~\cite{sun2021rangesparsenet} & \xmark & 63.0 & 62.6 & \bf 65.4 & \bf 60.7 \\
\hline
Deformable DETR \cite{zhu2020deformable} & \cmark & 59.6 & 59.2 & 45.8 & 36.2 \\
\hline
\textbf{\ours-3D} & \cmark & 63.9 & 63.7 & 61.5 & 53.7 \\
\end{tabular}
}
% }
%
\vspace{-0.8em}
%\medskip
\caption{\textbf{Comparison of \ours-3D in 3D object detection} on the Waymo Open \val set (LEVEL\_2 difficulty). Despite the lack of any class-specific optimization, \ours-3D is surprisingly effective and even competitive on the Vehicle category.}
\vspace{-0.1in}
\label{tab:3d_obj}
\end{table}

%% file: supplementary.tex
%%%%%%%%% BODY TEXT
\section{Additional Results}

\begin{table*}
\centering
\footnotesize
{

\tablestyle{6pt}{1.1}
\begin{tabular}{l|c|c|c|c|c|c|c|c|c|c|c}
 & AP$\uparrow$ & $\text{AP}_{\text{S}}\!\!\uparrow$ & $\text{AP}_{\text{M}}\!\!\uparrow$ & $\text{AP}_{\text{L}}\!\!\uparrow$ & AP$^\text{m}\!\!\uparrow$ & $\text{AP}^\text{m}_{\text{S}}\!\!\uparrow$ & $\text{AP}^\text{m}_{\text{M}}\!\!\uparrow$ & $\text{AP}^\text{m}_{\text{L}}\!\!\uparrow$ & \# params & FLOPs & fps\\
\shline
\boxattn {\it \ only} & 48.7 & 31.6 & 52.3 & 63.2 & - & - & - & - & 39.7M & 167G & 17.3 \\
w/ dropout & 47.8 & 30.9 & 50.6 & 62.1 & - & - & - & - & 39.7M & 167G & 17.3 \\
\hline
\instanceattn & 50.0 & 32.4 & 53.3 & 64.5 & 42.7 & 22.7 & 45.9 & 61.5 & 40.1M & 240G & 12.5 \\
\end{tabular}

}
\vspace{-0.8em}
\caption{\textbf{More \ours-2D ablation} on the COCO 2017 \val set using a R-50 backbone pretrained on ImageNet. By removing dropout during training, our \ours-2D shows better performance in all metrics without any extra computation. Note that we report the inference speed of our network including the post-processing time.}
\label{tab:sup_ablation}
\vspace{-0.1in}
\end{table*}

Here, we provide a comparison on the effects of dropout during the training of \ours. Specifically, we evaluate \ours-2D with and without dropout in Table \ref{tab:sup_ablation} along with the speed (fps), \# params, and FLOPs computation of our architecture.

\section{Prediction Head in \ours}

In order to take advantage of spatial information from the box-attention module, we design the prediction head to regress relative offsets \wrt the corresponding reference window. The reference window is used as the initial guess of the bounding box in both object proposal and detection stage. We denote $\sigma$ and $\sigma^{-1}$ as the $\mathop{\mathrm{sigmoid}}$ and the $\mathop{\mathrm{inverse\ sigmoid}}$ function, respectively. This design allows the prediction head of \ours to utilize the information of features in box-attention which learn to predict offsets from a reference window. Note that, we use $0 {=} \sigma^{-1}(0.5)$ for the height prediction of the 3D bounding boxes since we have no information in the reference window.

\begin{figure*}[ht]
\centering
\includegraphics[width=1\linewidth]{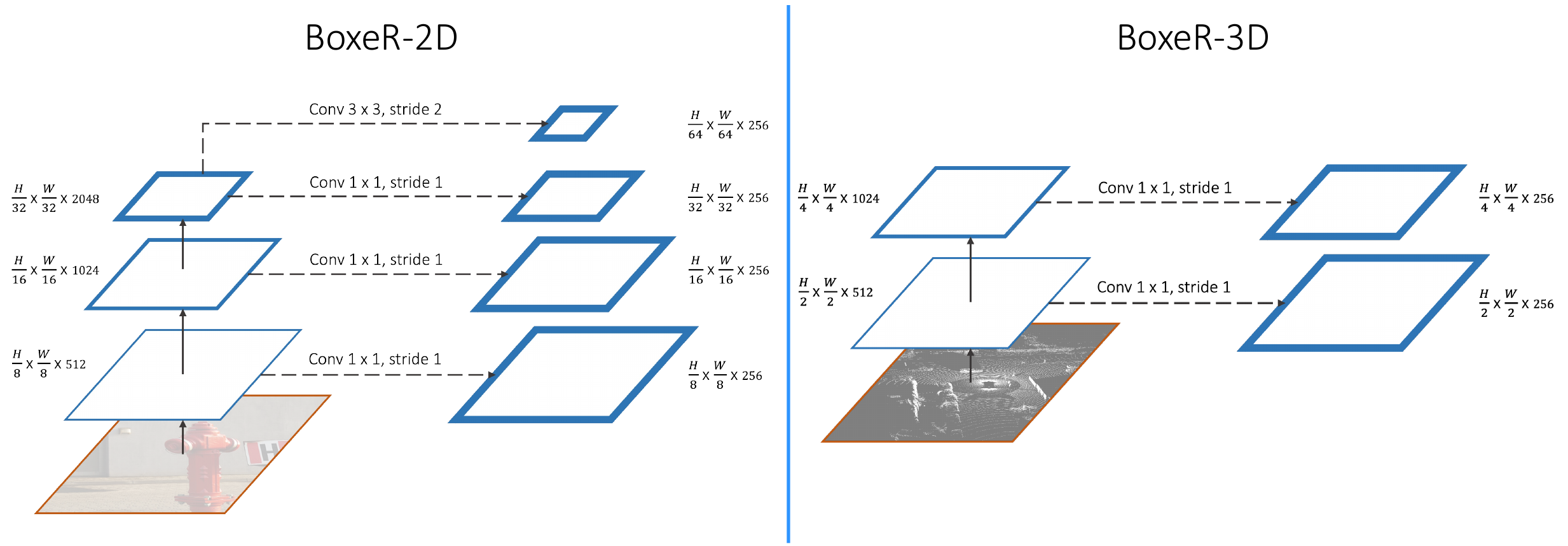}\\[-8pt]
\caption{\textbf{Multi-scale feature maps} for \ours-2D (left) and \ours-3D (right).}
\label{fig:ms}
\end{figure*}

\boldparagraph{\ours-2D.} Given a pre-defined reference window $b_q {=} [x, y, w_x, w_y] \in [0, 1]^4$, the predicted bounding box $\hat{b}_q {=} \Big[\sigma \big(\Delta_x {+} \sigma^{-1}(x)\big), \sigma \big(\Delta_y {+} \sigma^{-1}(y)\big), \sigma \big(\Delta_{w_x} {+} \sigma^{-1}(w_x)\big), 
\\ 
\sigma \big(\Delta_{w_y} {+} \sigma^{-1}(w_y)\big) \Big]$ where $\Delta_x$, $\Delta_y$, $\Delta_{w_x}$, and $\Delta_{w_y}$ are offsets of the bounding box center, width, and height predicted by the prediction head.

In the object proposal stage, the prediction head contains a 3-layer perceptron with a ReLU activation function and a hidden dimension $d$ for the offsets prediction and a linear projection for the bounding box binary classification (\ie, foreground and background). The encoder features of the top scoring proposals are picked as object queries in the decoder, while its bounding boxes serve as its reference windows (the object queries and encoder features do not have gradient flow).

In the prediction stage, the same architecture is used to predict the offsets from the reference windows and to classify object categories. The mask prediction head contains a deconv layer with $2 {\times} 2$ kernel size and stride $2$ followed by two $1 {\times} 1$ conv layers with ReLU activation function and hidden dimension $d$. Similar to \cite{he2017maskrcnn}, the last conv layer outputs a mask prediction of $28 {\times} 28 {\times} \mathop{\mathrm{num\_class}}$ for each bounding box.

\boldparagraph{\ours-3D.} Given a reference window $b_q {=} [x, y, w_x, w_y, \theta] \in [0, 1]^5$, the predicted bounding box $\hat{b}_q {=} \Big[\sigma \big(\Delta_x {+} \sigma^{-1}(x)\big), \sigma \big(\Delta_y {+} \sigma^{-1}(y)\big), \sigma \big(\Delta_z\big), \sigma \big(\Delta_{w_x} {+} 
\\ 
\sigma^{-1}(w_x)\big), \sigma \big(\Delta_{w_y} {+} \sigma^{-1}(w_y)\big), \sigma \big(\Delta_{w_z}\big), \sigma \big(\Delta_{\theta} {+} \sigma^{-1}(\theta) \big) \Big]$ where $\Delta_x$, $\Delta_y$, $\Delta_z$, $\Delta_{w_x}$, $\Delta_{w_y}$, $\Delta_{w_z}$, and $\Delta_{\theta}$ are the offsets of the bounding box center, length, width, height, and angle predicted by the prediction head. Note that, we normalize the center and size of the 3D bounding box by the detection range. The angle of the 3D bounding box is converted to the range $[0, 2\pi]$ and normalized by $2\pi$.

The prediction head in both the object proposal and prediction stage shares the same architecture: a 3-layer perceptron with a ReLU activation function and a hidden dimension $d$ for offsets prediction of the 3D bounding boxes, and a linear projection for the classification task (binary classification for object proposal and multi-class classification for prediction). Similarly, the prediction head is applied to each of the encoder features in the object proposal stage. Since \ours-3D uses reference windows of three angles per query, the prediction head outputs three object proposals each of which corresponds to a reference window of one angle. The top scoring proposals are picked in the same way as in \ours-2D.

\section{Losses in \ours Training}

We review the losses that are used during our training in Table \ref{tab:loss_weight}. \ours-2D is trained with a classification and a 2D box loss for the object proposal stage. In the prediction stage, classification, 2D box, and mask loss are used. The Hungarian matcher \cite{kuhn1955hungarian} in \ours-2D does not use the mask cost but only the classification and box cost as in \cite{zhu2020deformable}.

For 3D object detection, we train \ours-3D with a classification and 3D box loss for both the object proposal and prediction stage. Due to the limitation of our computational resources, we use the same loss weights as in \ours-2D except for the weight of the angle loss. Following \cite{nicolas2020detr}, the 3D Hungarian matcher is designed to be consistent with our setting of loss weights.

\section{Multi-scale feature maps for \ours}

The multi-scale feature maps are constructed for \ours-2D and \ours-3D as in Fig. \ref{fig:ms}. In end-to-end object detection and instance segmentation, we use ResNet \cite{kaiming2016resnet} and ResNeXt \cite{xie2017resnext} as backbones for \ours-2D. As discussed in Section 4, we follow \cite{zhu2020deformable} to construct our multi-scale feature maps. 

\begin{figure}
\centering
\includegraphics[width=0.3\linewidth]{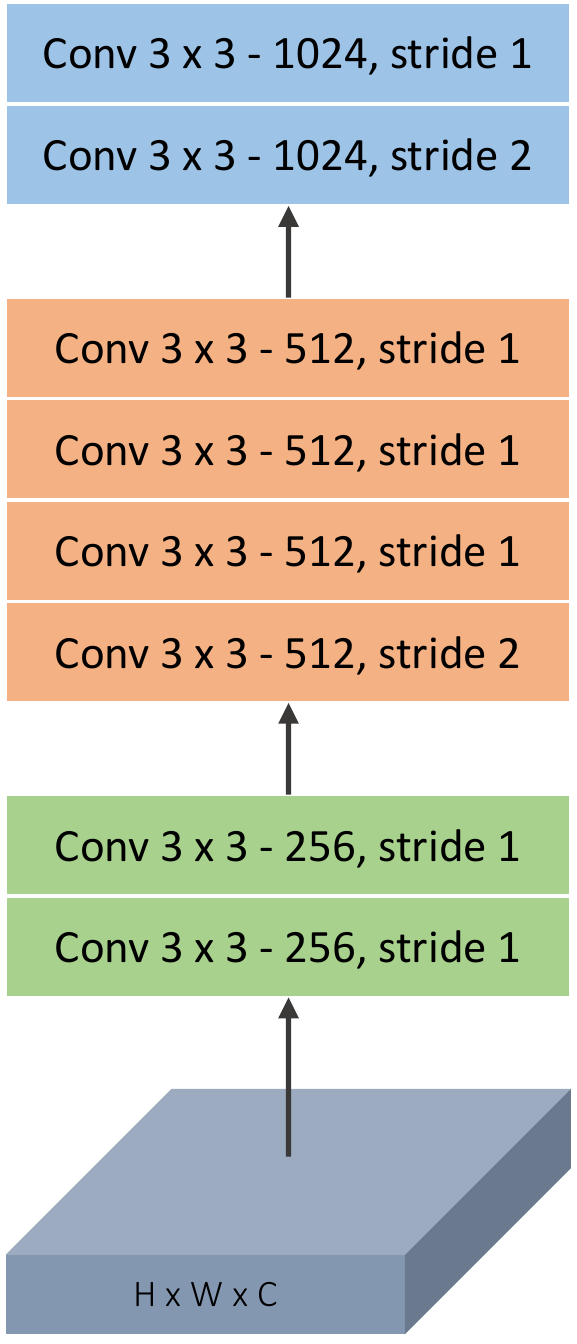}\\[-8pt]
\caption{\textbf{Convolutional backbone} for \ours-3D.}
\label{fig:backbone3d}
\vspace{-0.1in}
\end{figure}

In end-to-end 3D object detection, we first extract bird's-eye view image features from PointPillar \cite{lang2019pointpillar} with a grid size of $(0.32\text{m}, 0.32\text{m})$. The detection range is $[-75.0\text{m}, 75.0\text{m}]$ for the $x$ and $y$ axis, and $[-4\text{m}, 8\text{m}]$ for the $z$ axis. This results in a feature map $x \in \mathbb{R}^{H {\times} W {\times} C}$ where $H {=} W {=} 468$ and $C{=} 128$. In order to construct multi-scale feature maps, we use a convolutional backbone as in Fig. \ref{fig:backbone3d}. Note that each convolution layer is followed by a batch normalization layer and a ReLU activation layer.

%%%%%%%%%%%%%%%%%%%%%%%%%%%%%%%%%%%%%%%%%%%%%%%%%%%%%%%%%%%%%%%%%%%
\begin{table*}
\centering
\footnotesize
{
\begin{threeparttable}
\tablestyle{6pt}{1}
\begin{tabular}{l|cc|c|c|cc}
 & \multicolumn{2}{c|}{Box Loss} & Angle Loss & Classification Loss & \multicolumn{2}{c}{Mask Loss} \\
 & L1 loss & GIoU loss & L1 loss & Focal loss & BCE loss & DICE/F-1 loss \\
\shline
\ours-2D & 5 & 2 & n/a & 2 & 5 & 5 \\
\ours-3D & 5 & 2 & 4 & 2 & n/a & n/a \\
\end{tabular}
\end{threeparttable}
}
%\medskip
\vspace{-0.8em}
\caption{\textbf{Losses and weights} used in training of \ours-2D and \ours-3D (n/a: not available). We use the same loss weights between \ours-2D and \ours-3D.}
\label{tab:loss_weight}
\vspace{-0.2in}
\end{table*}
%%%%%%%%%%%%%%%%%%%%%%%%%%%%%%%%%%%%%%%%%%%%%%%%%%%%%%%%%%%%%%%%%%%%

\section{More Implementation Details}

\begin{figure}[t]
\centering

\begin{minipage}{0.45\textwidth}
\centering

\begin{minipage}[b]{.475\linewidth}
\includegraphics[width=\linewidth]{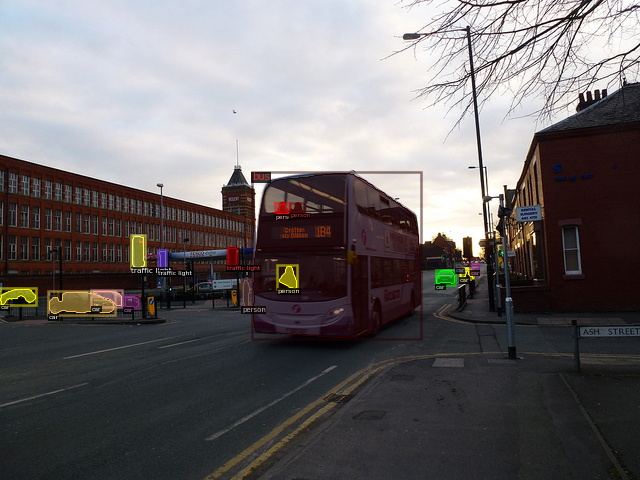}
\end{minipage}
\begin{minipage}[b]{.475\linewidth}
\includegraphics[width=\linewidth]{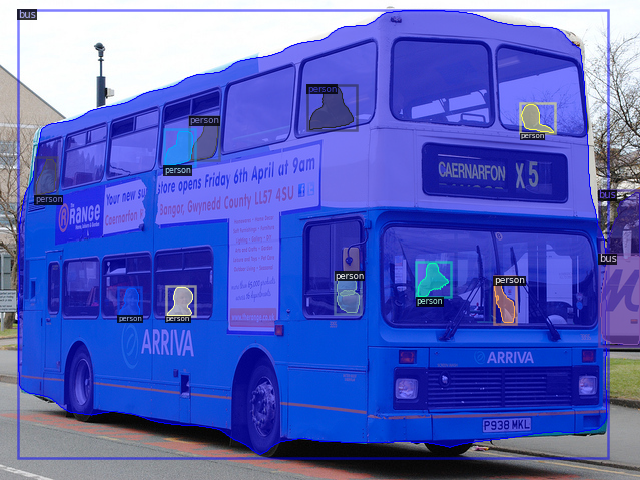}
\end{minipage}

% \begin{minipage}[b]{.475\linewidth}
% \includegraphics[width=\linewidth]{vis/vis_2d/val_700.png}
% \end{minipage}
% \begin{minipage}[b]{.475\linewidth}
% \includegraphics[width=\linewidth]{vis/vis_2d/val_990.png}
% \end{minipage}

\begin{minipage}[b]{.475\linewidth}
\includegraphics[width=\linewidth]{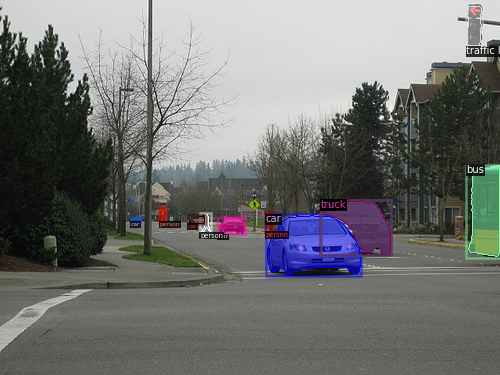}
\end{minipage}
\begin{minipage}[b]{.475\linewidth}
\includegraphics[width=\linewidth]{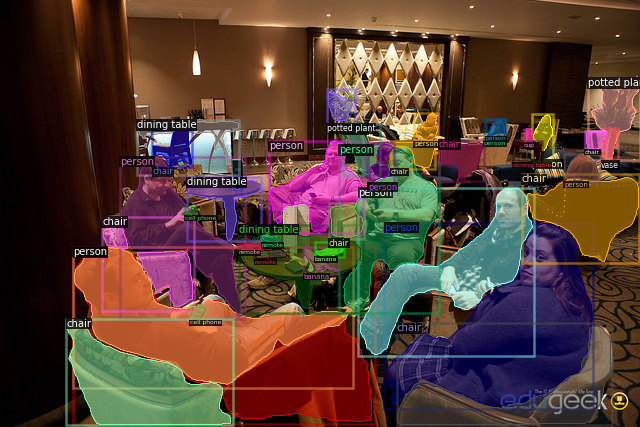}
\end{minipage}

\begin{minipage}[b]{.475\linewidth}
\includegraphics[width=\linewidth]{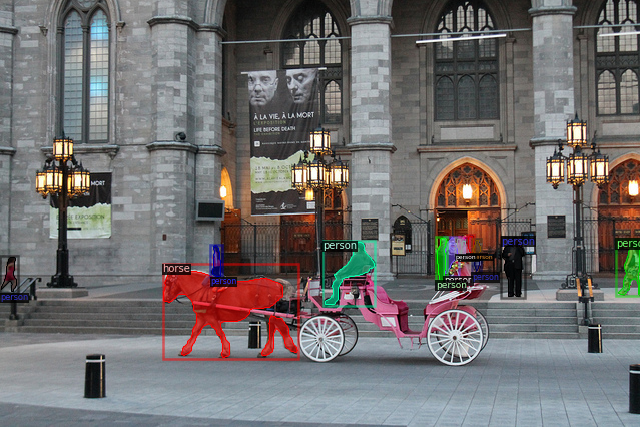}
\end{minipage}
\begin{minipage}[b]{.475\linewidth}
\includegraphics[width=\linewidth]{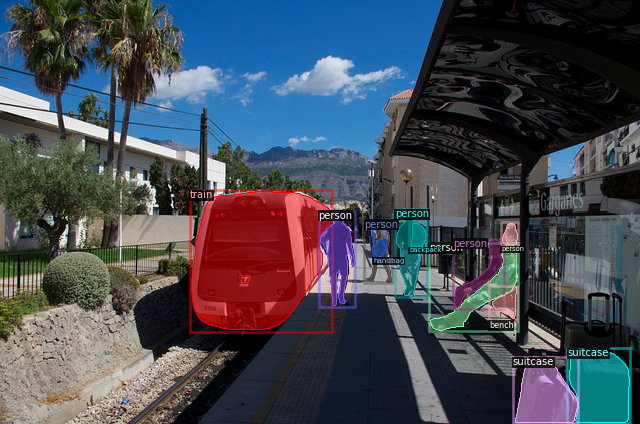}
\end{minipage}

% \begin{minipage}[b]{.45\linewidth}
% \includegraphics[width=\linewidth]{vis/vis_2d/val_906.png}
% \end{minipage}
% \begin{minipage}[b]{.45\linewidth}
% \includegraphics[width=\linewidth]{vis/vis_2d/val_782.png}
% \end{minipage}

\begin{minipage}[b]{.475\linewidth}
\includegraphics[width=\linewidth]{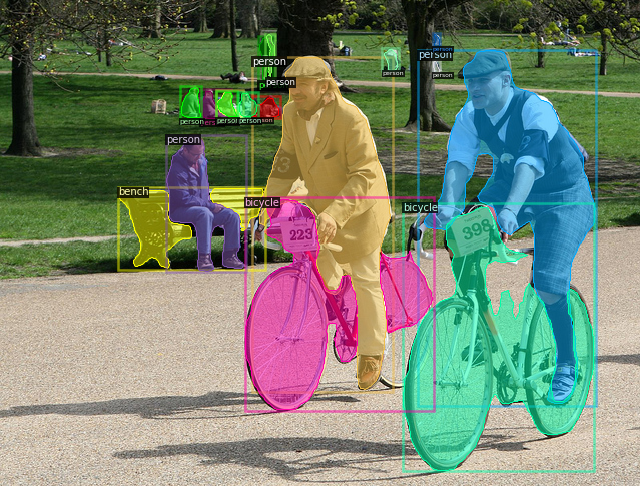}
\end{minipage}
\begin{minipage}[b]{.475\linewidth}
\includegraphics[width=\linewidth]{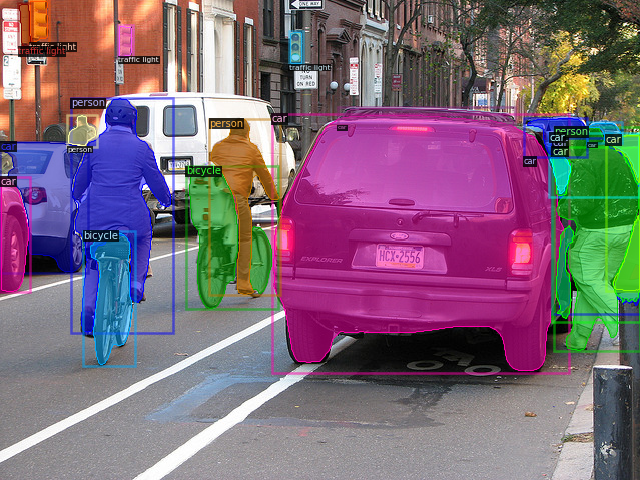}
\end{minipage}

\begin{minipage}[b]{.475\linewidth}
\includegraphics[width=\linewidth]{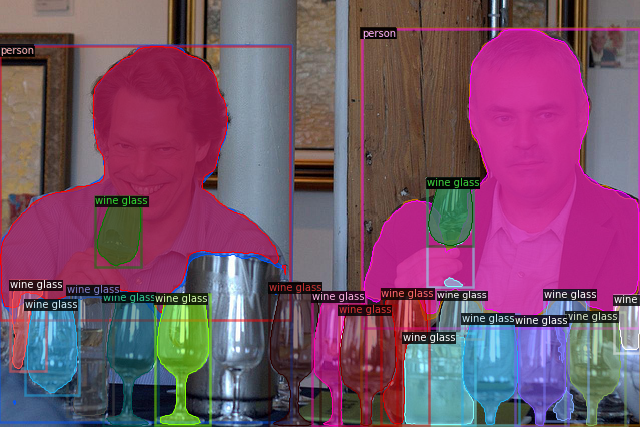}
\end{minipage}
\begin{minipage}[b]{.475\linewidth}
\includegraphics[width=\linewidth]{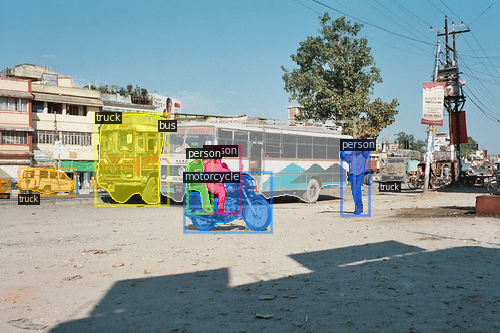}
\end{minipage}

% \begin{minipage}[b]{.475\linewidth}
% \includegraphics[width=\linewidth]{vis/vis_2d/val_725.png}
% \end{minipage}
% \begin{minipage}[b]{.475\linewidth}
% \includegraphics[width=\linewidth]{vis/vis_2d/val_906.png}
% \end{minipage}

\end{minipage}

\vspace{-0.8em}
\caption{\textbf{Qualitative results} for object detection and instance segmentation generated by \ours-2D in the COCO 2017 \test set.}
\label{fig:vis_2d_sup}
\vspace{-0.2in}
\end{figure}

\boldparagraph{Initialization of multi-scale box-attention.} Unlike deformable attention \cite{zhu2020deformable}, our box-attention module does not require a complicated initialization. Weight and bias parameters of the linear projection for generating attention weights are initialized to zero. For predicting the offsets of the reference windows, weight and bias parameters are initialized to zeros and randomly, respectively. By initializing uniform attention weights, box-attention gathers all information within the attended region to make its decision. Other parameters are randomly initialized.

\boldparagraph{Initialization of prediction head.} Following the initialization of the box-attention module, we initialize weight and bias parameters of the last layer in the 3-layer perceptron for offsets prediction to zero. Other parameters are randomly initialized. By doing so, the prediction head treats reference windows as its initial guess which is consistent with the box-attention module.

\boldparagraph{Spatial Encoding.} Each query $q \in \mathbb{R}^d$ in image features is assigned a reference window $b_{q} {=} [x, y, w_x, w_y]$ where $x, y$ are coordinates of the window center corresponding to the query position and $w_x, w_y$ are width and height of the reference window. As in \cite{nicolas2020detr}, we use a fixed absolute encoding to represent $\{x, y\}$ which adopts a sinusoid encoding to the 2D case. Specifically, $x$ and $y$ are independently encoded in $\frac{d}{2}$ features using sine and cosine functions with different frequencies. The $d$ channel position encoding is their concatenation. Similarly, we get a $d$ channel size encoding of $\{w_x, w_y\}$. The final spatial encoding of query $q$ is the summation of both position and size encodings.

\boldparagraph{Hyperparameter settings in \ours.} Within its box of interest, the box-attention module samples a grid of $2 {\times} 2$ ($m {=} 2$) while the instance-attention module samples a grid of $14 {\times} 14$ ($m {=} 14$). We set the hidden dim of \ours, $d {=} 256$; the hidden dim of feed-forward sub-layers, $d_\text{feed-forward} {=} 1024$; the number of attention heads, $l {=} 8$. This applies to both \ours-2D and \ours-3D. \ours-2D contains 6 encoder and decoder layers ($S {=} 6$) while \ours-3D consists of 2 encoder and decoder layers ($S {=} 2$).

During inference of \ours-2D, given \ours-2D decoder of $S$ decoder layers, we only output $x^\text{mask}_{S} \in \mathbb{R}^{N \times m \times m \times d}$ in the last decoder layer to speed up its prediction time.

\boldparagraph{Details of the transformation functions.} Our translation and scaling functions predict offsets $\Delta_x, \Delta_y, \Delta_{w_x}, \Delta_{w_y}$ \wrt the reference window $b_{q} {=} [x, y, w_x, w_y] \in [0, 1]^4$ of query $q$ using a linear projection on $q$ as below

\begin{equation}
\Delta_x = (qW_x^{\top} + b_x) * \frac{w_x}{\tau} \enspace,
\end{equation}
\begin{equation}
\Delta_y = (qW_y^{\top} + b_y) * \frac{w_y}{\tau} \enspace,
\end{equation}
\begin{equation}
\Delta_{w_x} = \mathop{\mathrm{max}}(qW_{w_x}^{\top} + b_{w_x}, 0) * \frac{w_x}{\tau} \enspace,
\end{equation}
\begin{equation}
\Delta_{w_y} = \mathop{\mathrm{max}}(qW_{w_y}^{\top} + b_{w_y}, 0) * \frac{w_y}{\tau} \enspace,
\end{equation}
where $W$ and $b$ are weights and biases of the linear projections; $\tau {=} 8$ is the temperature hyperparameter. The multiplication of the window size $\{w_x, w_y\}$ helps the prediction of the linear projection to be scale-invariant. For angle prediction in 3D object detection, the offset is predicted as
\begin{equation}
\Delta_{\theta} = (qW_{\theta}^{\top} + b_{\theta}) * \frac{1}{\tau_{\theta}} \enspace.
\end{equation}

\begin{figure}
\centering

\begin{minipage}{.475\textwidth}
\centering

\begin{minipage}[b]{.475\linewidth}
\includegraphics[width=\linewidth]{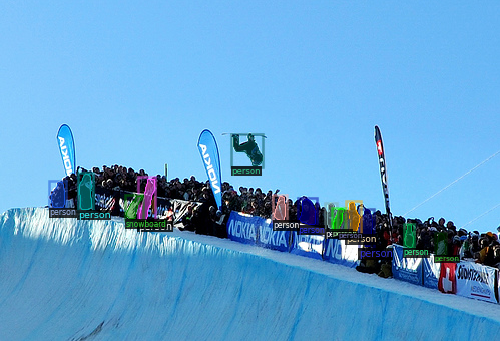}
\end{minipage}
\begin{minipage}[b]{.475\linewidth}
\includegraphics[width=\linewidth]{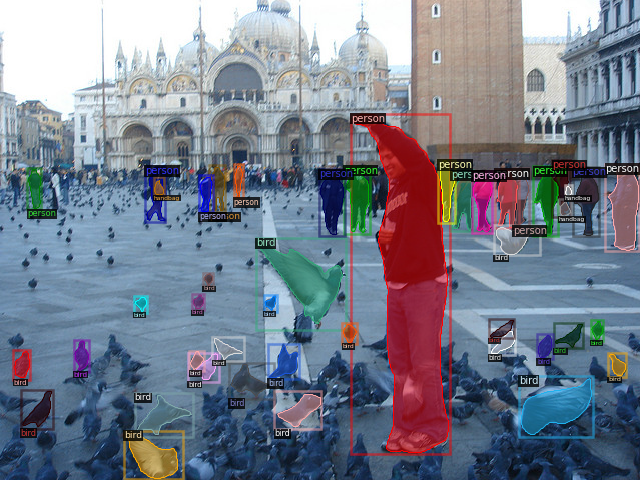}
\end{minipage}

\begin{minipage}[b]{.475\linewidth}
\includegraphics[width=\linewidth]{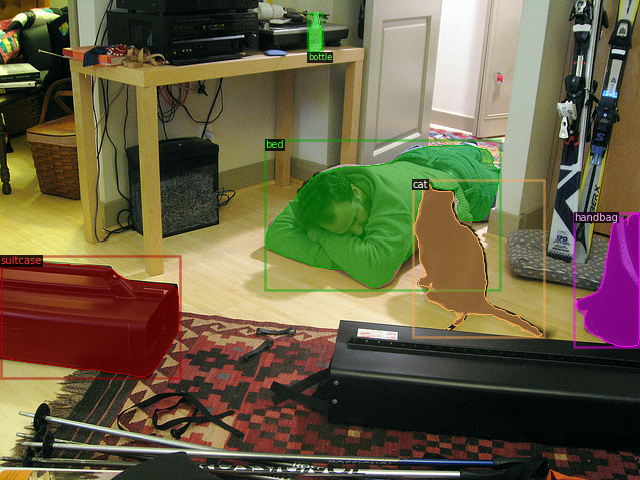}
\end{minipage}

\end{minipage}

\vspace{-0.8em}
\caption{\textbf{Failure cases of \ours-2D.} \ours-2D fails to predict very small objects in low light conditions. The last image shows that \ours-2D is able to predict the object bounding box and mask but fails to classify the category.}
\label{fig:failure_2d_sup}
\vspace{-0.2in}
\end{figure}

\boldparagraph{Implementation details of \ours-3D.} During training of \ours-3D, following \cite{shi2020pvrcnn} we use random flipping along the $x$ and $y$ axis, global scaling with a random factor sampled from $[0.95, 1.05]$, global rotation around the $z$ axis with a random angle sampled from $[-\frac{\pi}{4}, \frac{\pi}{4}]$. The ground truth sampling augmentation is also conducted to randomly ``paste'' ground-truth objects from other frames to the current one. Similar to the \ours-2D, we take the top-300 scoring encoder features as object queries along with the corresponding predicted bounding boxes as their reference window for the \ours-3D decoder. In the final prediction, we pick 125 predictions with the highest scores.

\section{More Qualitative Results}

\boldparagraph{2D object detection and instance segmentation.} We show extra qualitative results for object detection and instance segmentation of the \ours-2D with a R-101 backbone in Fig. \ref{fig:vis_2d_sup} and Fig. \ref{fig:failure_2d_sup}.

\begin{figure*}
\centering

\begin{minipage}{.475\textwidth}
\centering

\begin{minipage}[b]{.475\linewidth}
\includegraphics[width=\linewidth]{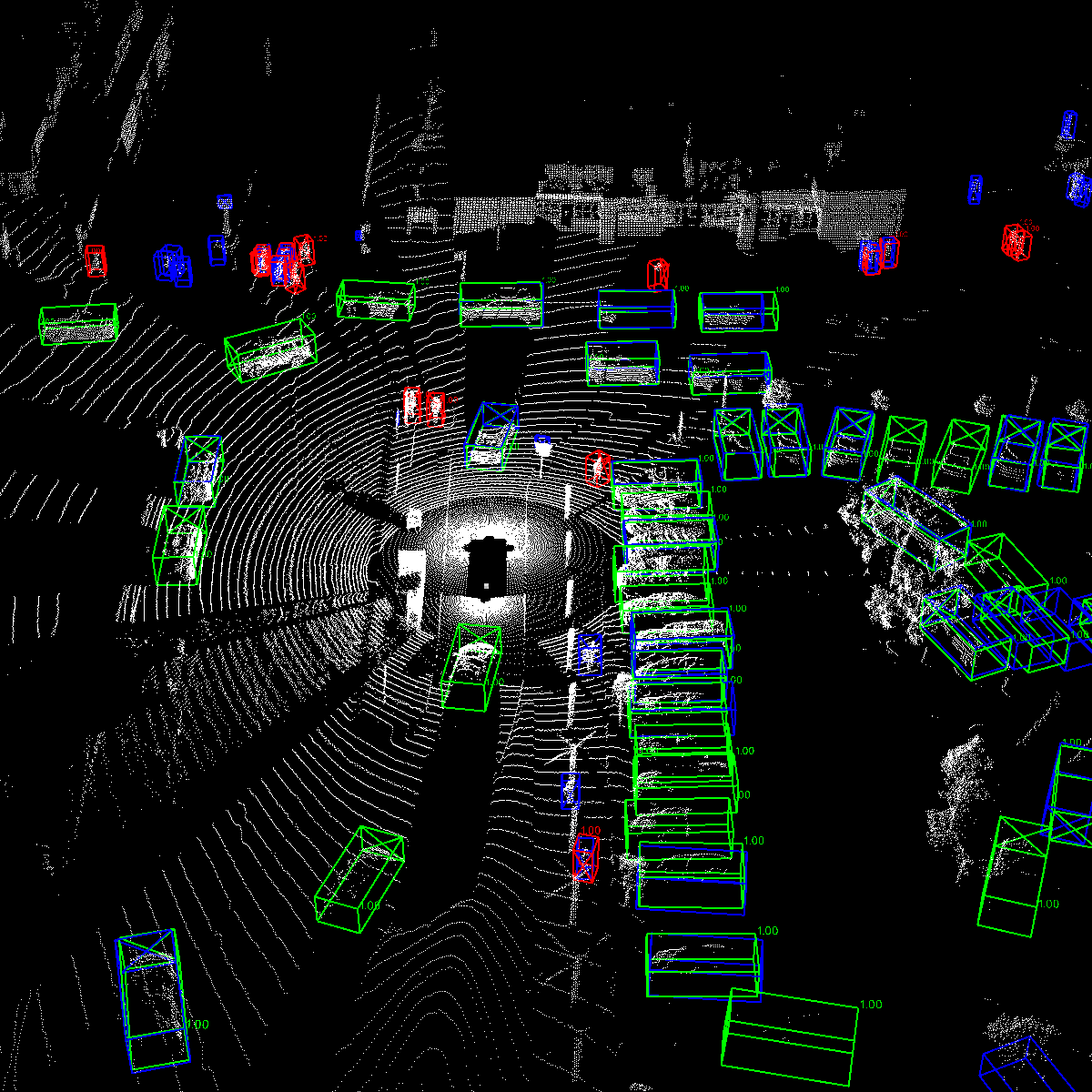}
\end{minipage}
\begin{minipage}[b]{.475\linewidth}
\includegraphics[width=\linewidth]{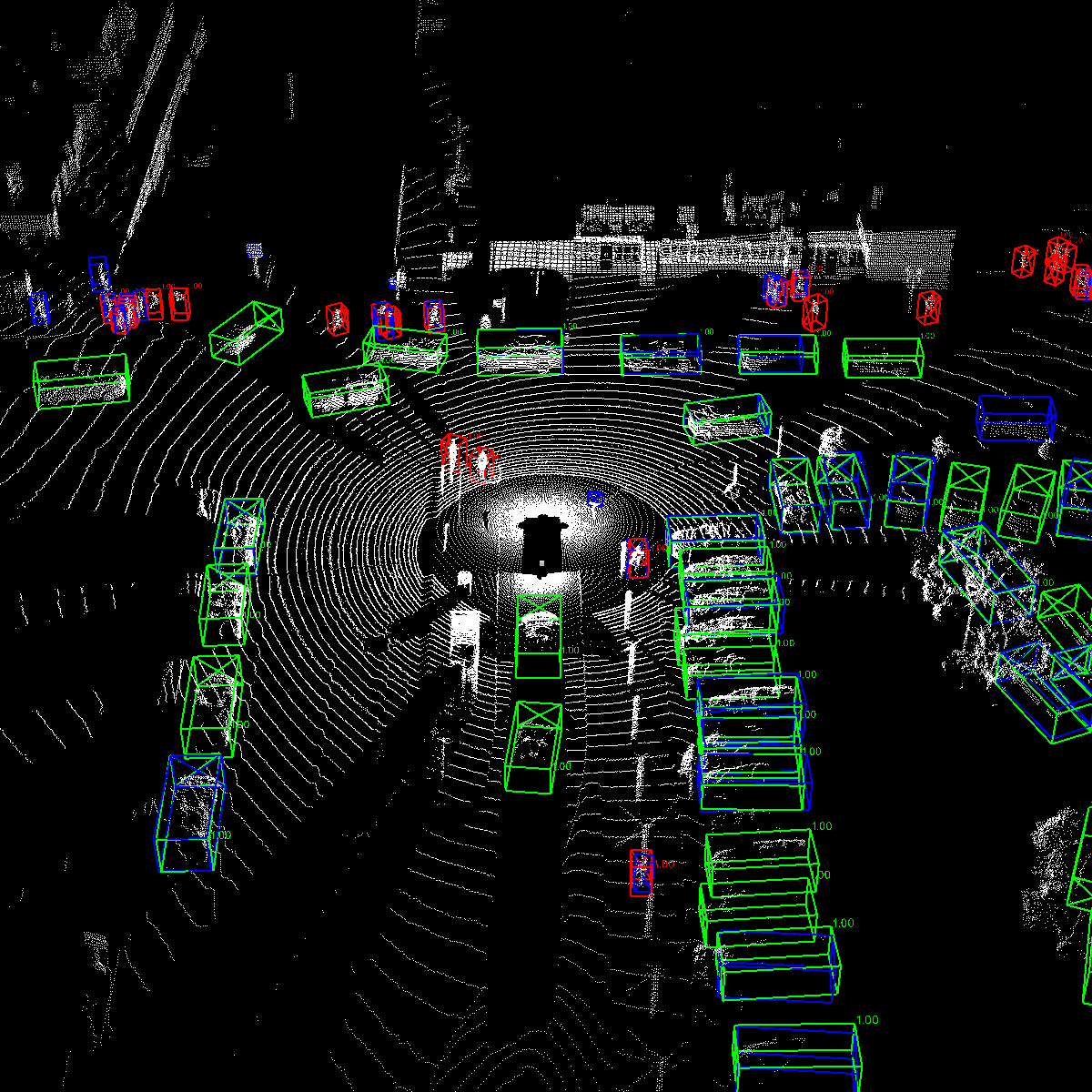}
\end{minipage}
\begin{minipage}[b]{.475\linewidth}
\includegraphics[width=\linewidth]{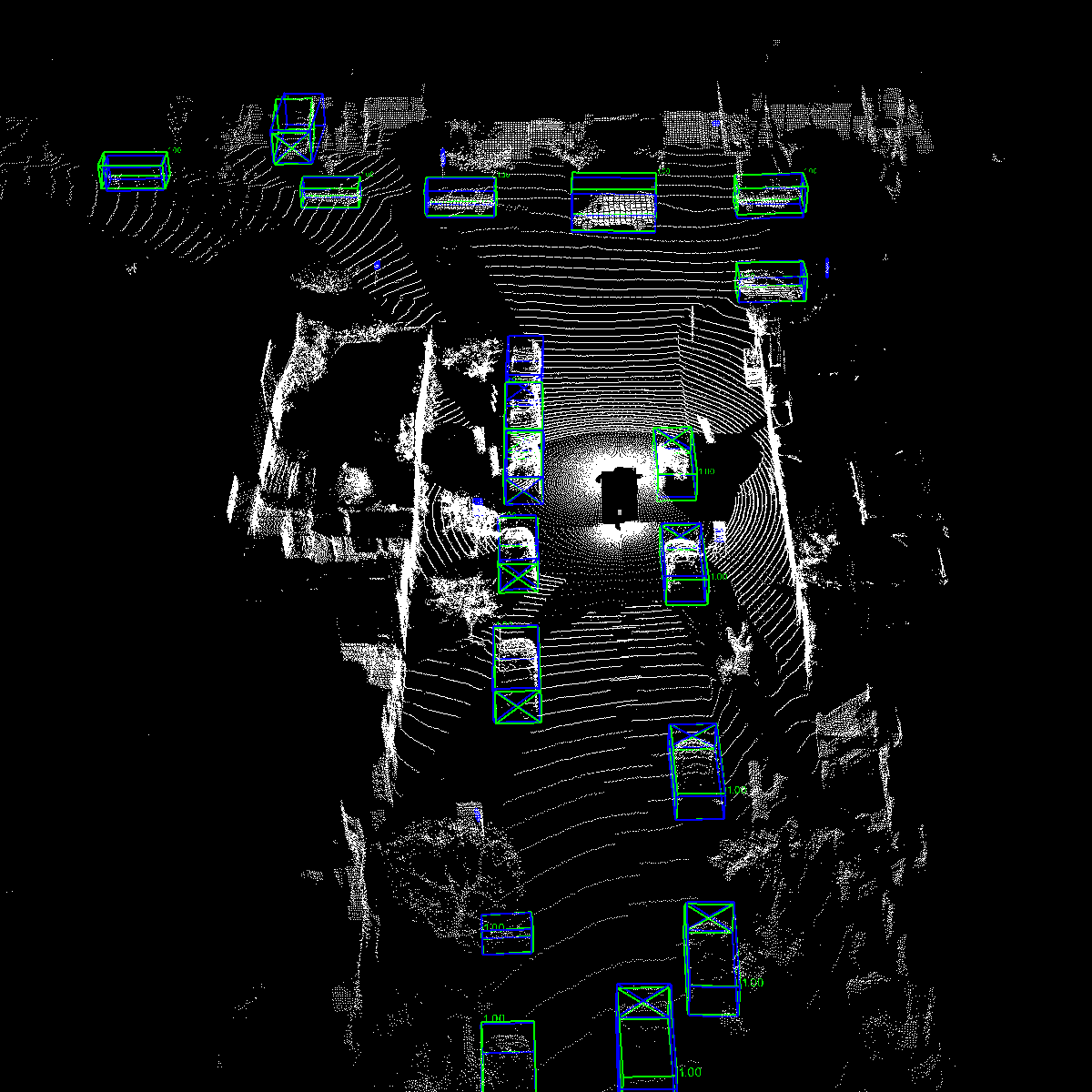}
\end{minipage}
\begin{minipage}[b]{.475\linewidth}
\includegraphics[width=\linewidth]{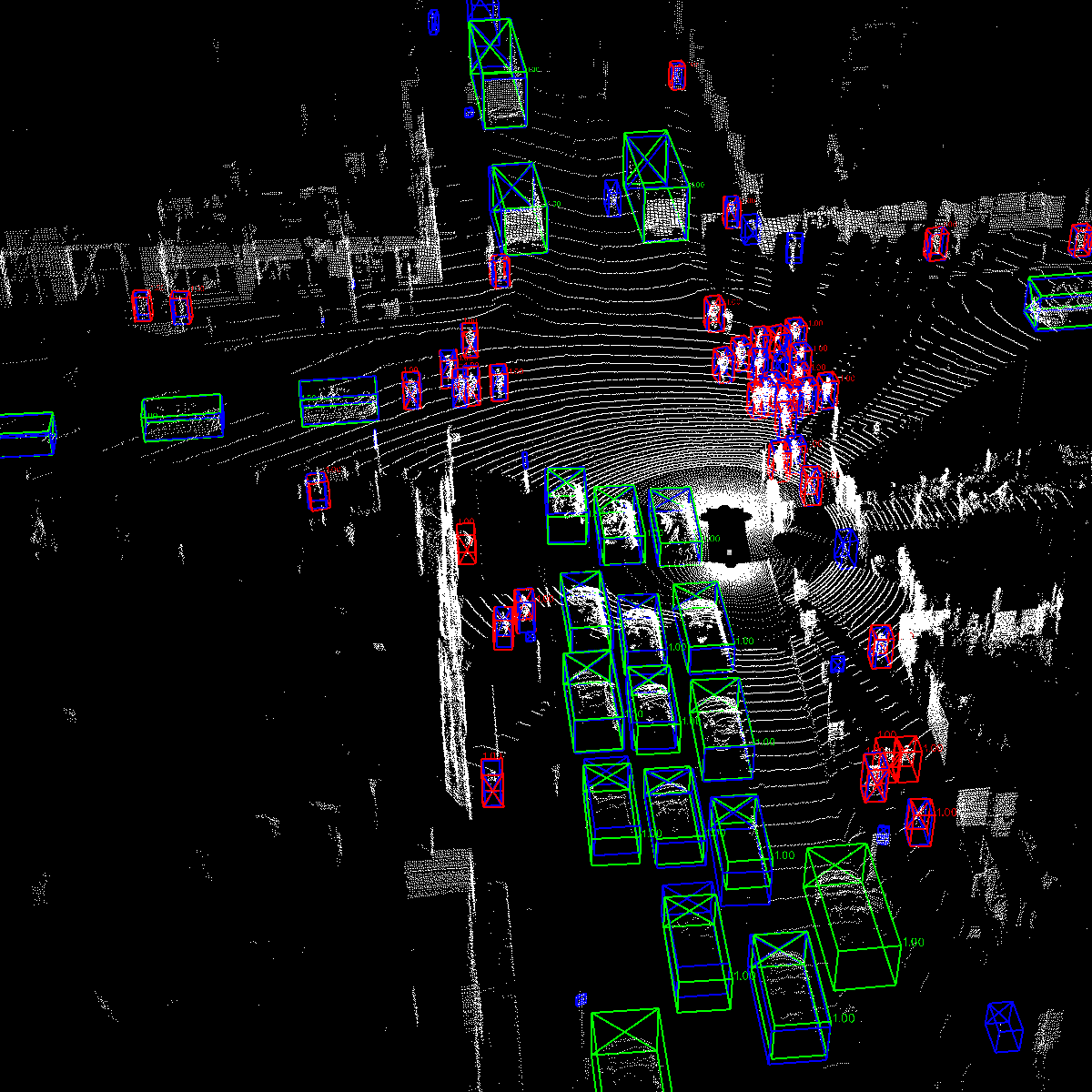}
\end{minipage}

\end{minipage}
\begin{minipage}{.475\textwidth}
\centering

\begin{minipage}[b]{.475\linewidth}
\includegraphics[width=\linewidth]{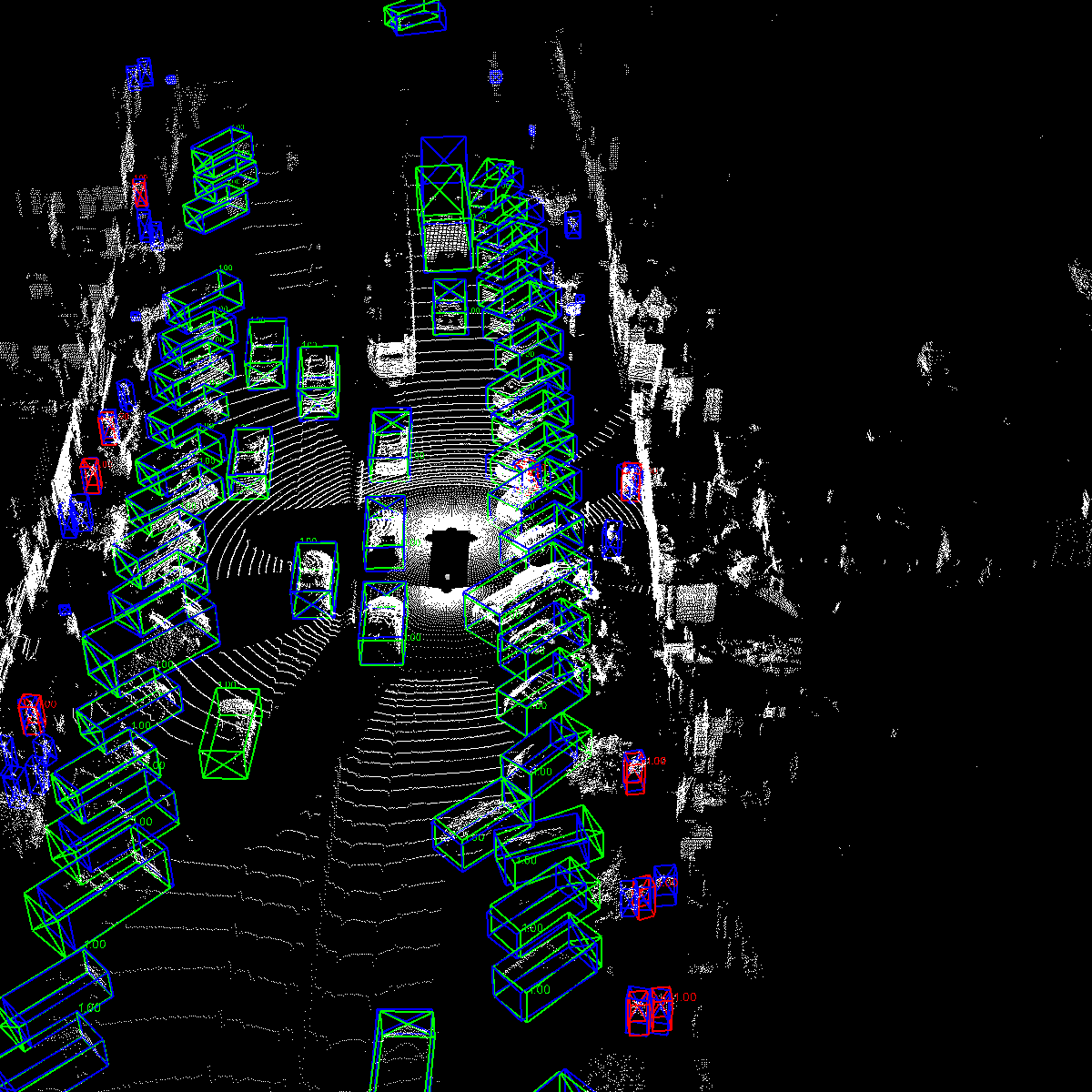}
\end{minipage}
\begin{minipage}[b]{.475\linewidth}
\includegraphics[width=\linewidth]{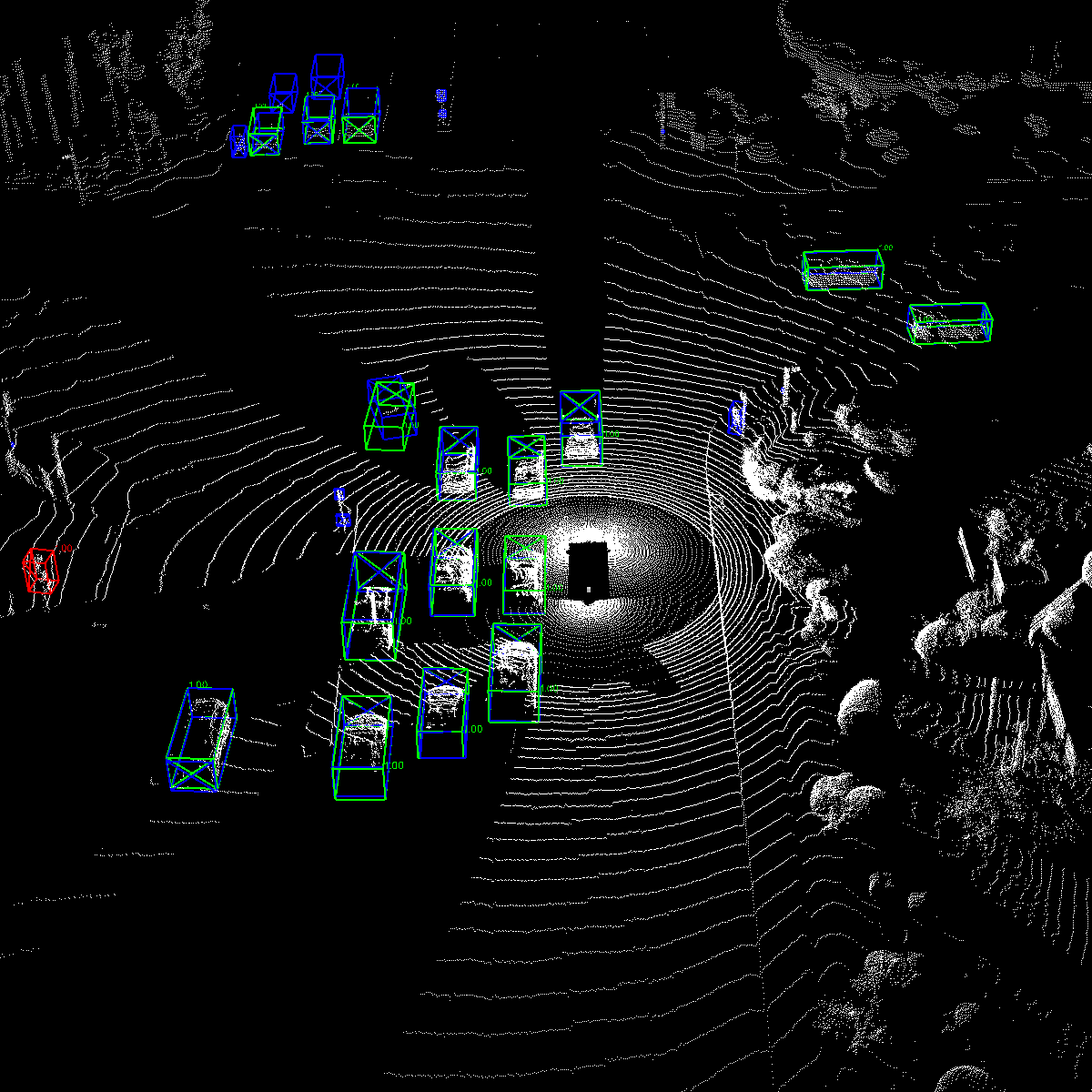}
\end{minipage}
\begin{minipage}[b]{.475\linewidth}
\includegraphics[width=\linewidth]{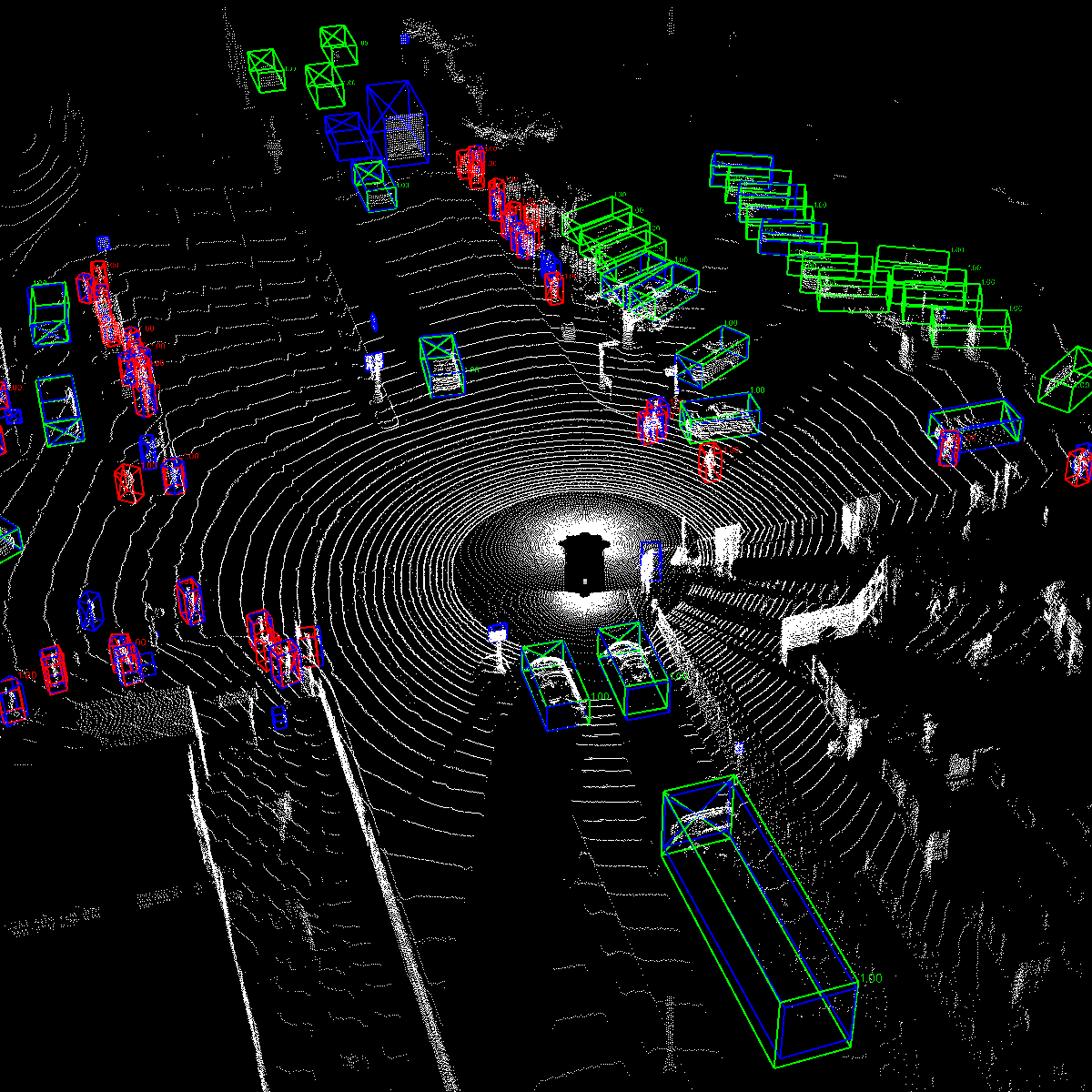}
\end{minipage}
\begin{minipage}[b]{.475\linewidth}
\includegraphics[width=\linewidth]{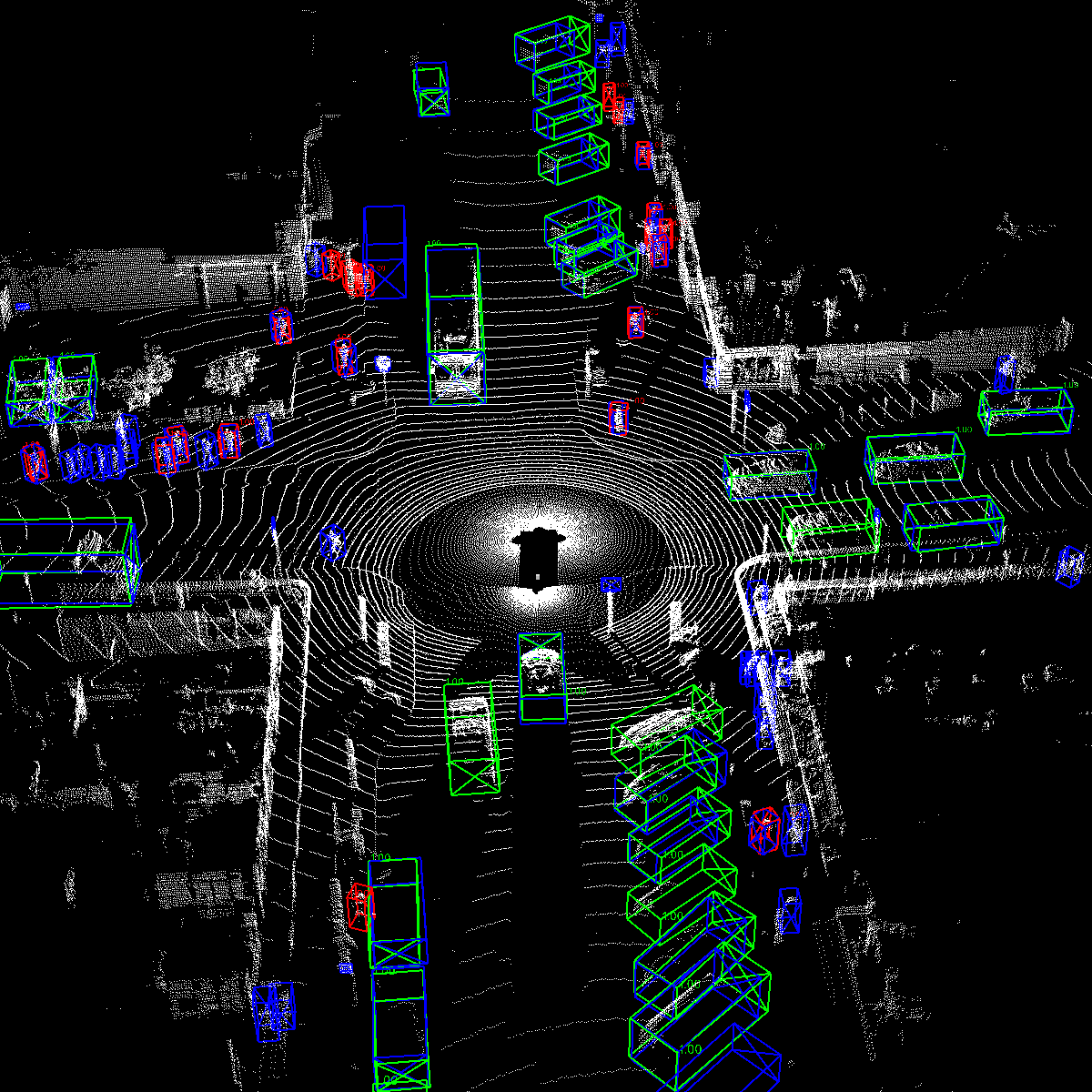}
\end{minipage}

\end{minipage}

\caption{\textbf{Qualitative results} for 3D object detection generated by \ours-3D on the Waymo \val set.}
\label{fig:vis_3d_sup}
\end{figure*}

\boldparagraph{3D object detection.} We show extra qualitative results for 3D object detection of the \ours-3D in Fig. \ref{fig:vis_3d_sup}. The ground-truth boxes are denoted in blue while the predicted pedestrian and vehicle are in red and green. It can be seen that \ours-3D gives a high accuracy for vehicle prediction. Note that, \ours-3D detects some vehicles that are missed by annotators. However, \ours-3D still struggles with detecting all pedestrians.